\documentclass[sigconf]{acmart}

\usepackage{booktabs} 

\usepackage{booktabs} 
\usepackage{algorithm}
\usepackage{algorithmic}
\usepackage{color}
\usepackage{amsmath}
\usepackage{amssymb}
\usepackage{subfigure}
\usepackage{graphicx}
\usepackage{epstopdf}
\usepackage{multirow}
\usepackage{booktabs} 

\usepackage{lineno}
\usepackage{multirow}

\newcommand{\bred}{\bf }
\newcommand{\snd}{\em }

\copyrightyear{2018}
\acmYear{2018}
\setcopyright{acmcopyright}
\acmConference[KDD 2018]{24th ACM SIGKDD International Conference on Knowledge Discovery \& Data Mining}{August 19--23, 2018}{London, United Kingdom}
\acmBooktitle{KDD 2018: 24th ACM SIGKDD International Conference on Knowledge Discovery \& Data Mining, August 19--23, 2018, London, United Kingdom}
\acmPrice{15.00}
\acmDOI{10.1145/3219819.3220060}
\acmISBN{978-1-4503-5552-0/18/08}

\begin{document}
\title{Multilevel Wavelet Decomposition Network for Interpretable Time Series Analysis}

\author{Jingyuan Wang, Ze Wang, Jianfeng Li, Junjie Wu}
\affiliation{%
	\institution{Beihang University, Beijing, China}
}\email{{jywang, ze.w,leejianfeng,wujj}@buaa.edu.cn}



\begin{abstract}
	Recent years have witnessed the unprecedented rising of time series from almost all kindes of academic and industrial fields. Various types of deep neural network models have been introduced to time series analysis, but the important frequency information is yet lack of effective modeling. In light of this, in this paper we propose a wavelet-based neural network structure called multilevel Wavelet Decomposition Network (mWDN) for building frequency-aware deep learning models for time series analysis. mWDN preserves the advantage of multilevel discrete wavelet decomposition in frequency learning while enables the fine-tuning of all parameters under a deep neural network framework. Based on mWDN, we further propose two deep learning models called Residual Classification Flow (RCF) and multi-frequecy Long Short-Term Memory (mLSTM) for time series classification and forecasting, respectively. The two models take all or partial mWDN decomposed sub-series in different frequencies as input, and resort to the back propagation algorithm to learn all the parameters globally, which enables seamless embedding of wavelet-based frequency analysis into deep learning frameworks. Extensive experiments on 40 UCR datasets and a real-world user volume dataset demonstrate the excellent performance of our time series models based on mWDN. In particular, we propose an importance analysis method to mWDN based models, which successfully identifies those time-series elements and mWDN layers that are crucially important to time series analysis. This indeed indicates the interpretability advantage of mWDN, and can be viewed as an indepth exploration to interpretable deep learning.
	
\end{abstract}

%
%



\maketitle

\section{Introduction}
A time series is a series of data points indexed in time order. Methods for time series analysis could be classified into two types: time-domain methods and frequency-domain methods.\footnote{\url{https://en.wikipedia.org/wiki/Time_series}} Time-domain methods consider a time series as a sequence of ordered points and analyze correlations among them. Frequency-domain methods use transform algorithms, such as discrete Fourier transform and Z-transform, to transform a time series into a frequency spectrum, which could be used as features to analyze the original series.

In recent years, with the booming of deep learning concept, various types of deep neural network models have been introduced to time series analysis and achieved state-of-the-art performances in many real-life applications~\cite{baselines,cardiologist}. Some well-known models include Recurrent Neural Networks (RNN)~\cite{rnn} and Long Short-Term Memory (LSTM)~\cite{lstm} that use memory nodes to model correlations of series points, and Convolutional Neural Network (CNN) that uses trainable convolution kernels to model local shape patterns~\cite{qiliu}. Most of these models fall into the category of time-domain methods without leveraging frequency information of a time series, although some begin to consider in indirect ways~\cite{MCNN,cwrnn}.



Wavelet decompositions~\cite{wavelet} are well-known methods for capturing features of time series both in time and frequency domains. Intuitively, we can employ them as feature engineering tools for data preprocessing before a deep modeling. While this loose coupling way might improve the performance of raw neural network models~\cite{waveletnn}, they are not globally optimized with independent parameter inference processes. How to integrate wavelet transforms into the framework of deep learning models remains a great challenge.

In this paper, we propose a wavelet-based neural network structure, named {\it multilevel Wavelet Decomposition Network} (mWDN), to build frequency-aware deep learning models for time series analysis. Similar to the standard {\it Multilevel Discrete Wavelet Decomposition} (MDWD) model~\cite{mallat1989theory}, mWDN can decompose a time series into a group of sub-series with frequencies ranked from high to low, which is crucial for capturing frequency factors for deep learning. Different from MDWD with fixed parameters, however, all parameters in mWDN can be fine-turned to fit training data of different learning tasks. In other words, mWDN can take advantages of both wavelet based time series decomposition and the learning ability of deep neural networks.

Based on mWDN, two deep learning models, {\it i.e.}, Residual Classification Flow (RCF) and multi-frequency Long Short-Term Memory (mLSTM), are designed for time series classification (TSC) and forecasting (TSF), respectively. The key issue in TSC is to extract as many as possible representative features from time series. The RCF model therefore adopts the mWDN decomposed results in different levels as inputs, and employs a pipelined classifier stack to exploit features hidden in sub-series through residual learning methods. For the TSF problem, the key issue turns to inferring future states of a time series according to the hidden trends in different frequencies. Therefore, the mLSTM model feeds all mWDN decomposed sub-series in high frequencies into independent LSTM models, and ensembles all LSTM outputs for final forecasting. Note that all parameters of RCF and mLSTM including the ones in mWDN are trained using the back propagation algorithm in an {\it end-to-end} manner. In this way, the wavelet-based frequency analysis is seamlessly embedded into deep learning frameworks.

We evaluate RCF on 40 UCR time series datasets for TSC, and mLSTM on a real-world user-volume time series dataset for TSF. The results demonstrate their superiorities to state-of-the-art baselines and the advantages of mWDN with trainable parameters. As a nice try for interpretable deep learning, we further propose an importance analysis method to mWDN based models, which successfully identifies those time-series elements and mWDN layers that are crucially important to the success of time series analysis. This indicates the interpretability advantage of mWDN by integrating wavelet decomposition for frequency factors.

\begin{figure}[t]\centering
	\subfigure[Illustration of the mWDN Framework]
	{\label{fig:mWDN_framework}\includegraphics[width=\columnwidth]{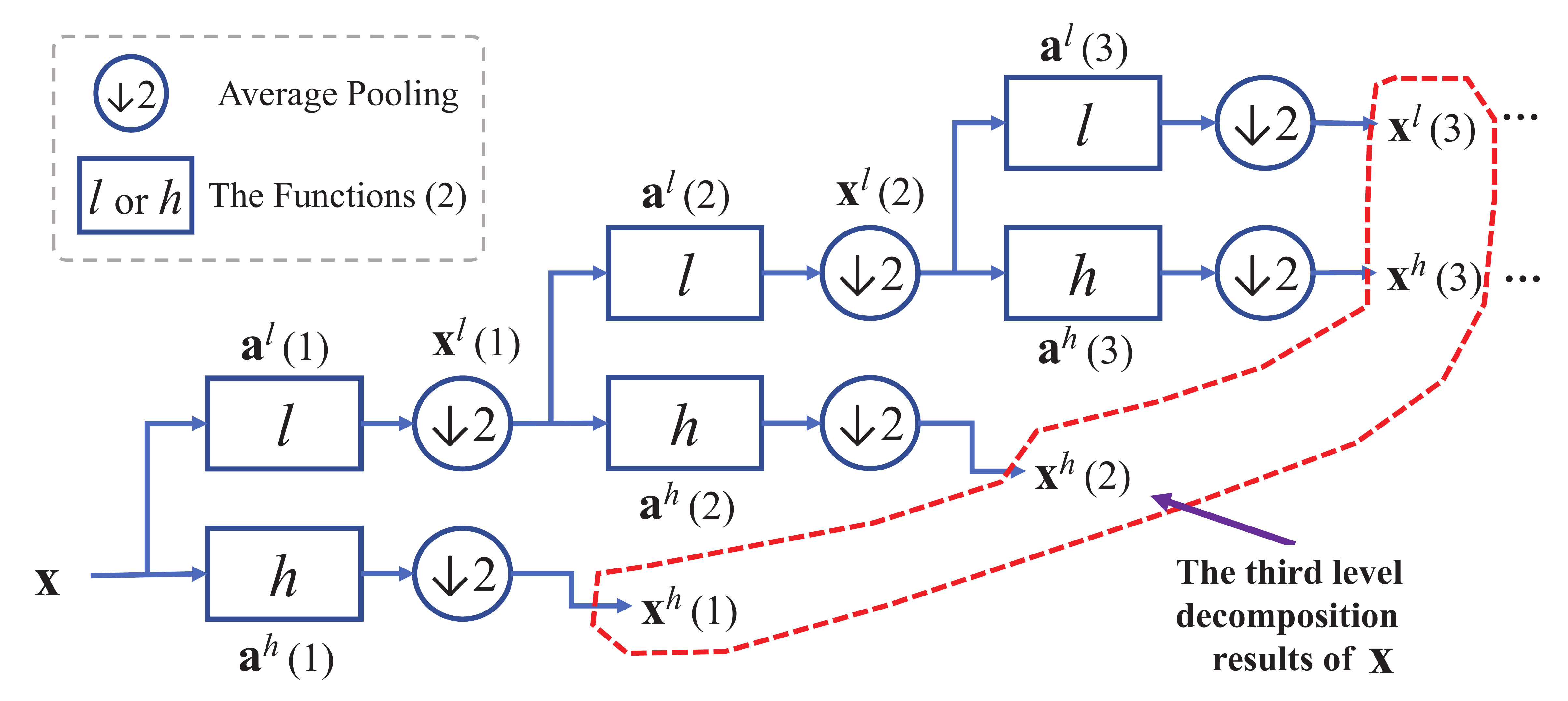}}\\[-0.2cm]
	\subfigure[Approximative Discrete Wavelet Transform]
	{\label{fig:mWDN_layer}\includegraphics[width=\columnwidth]{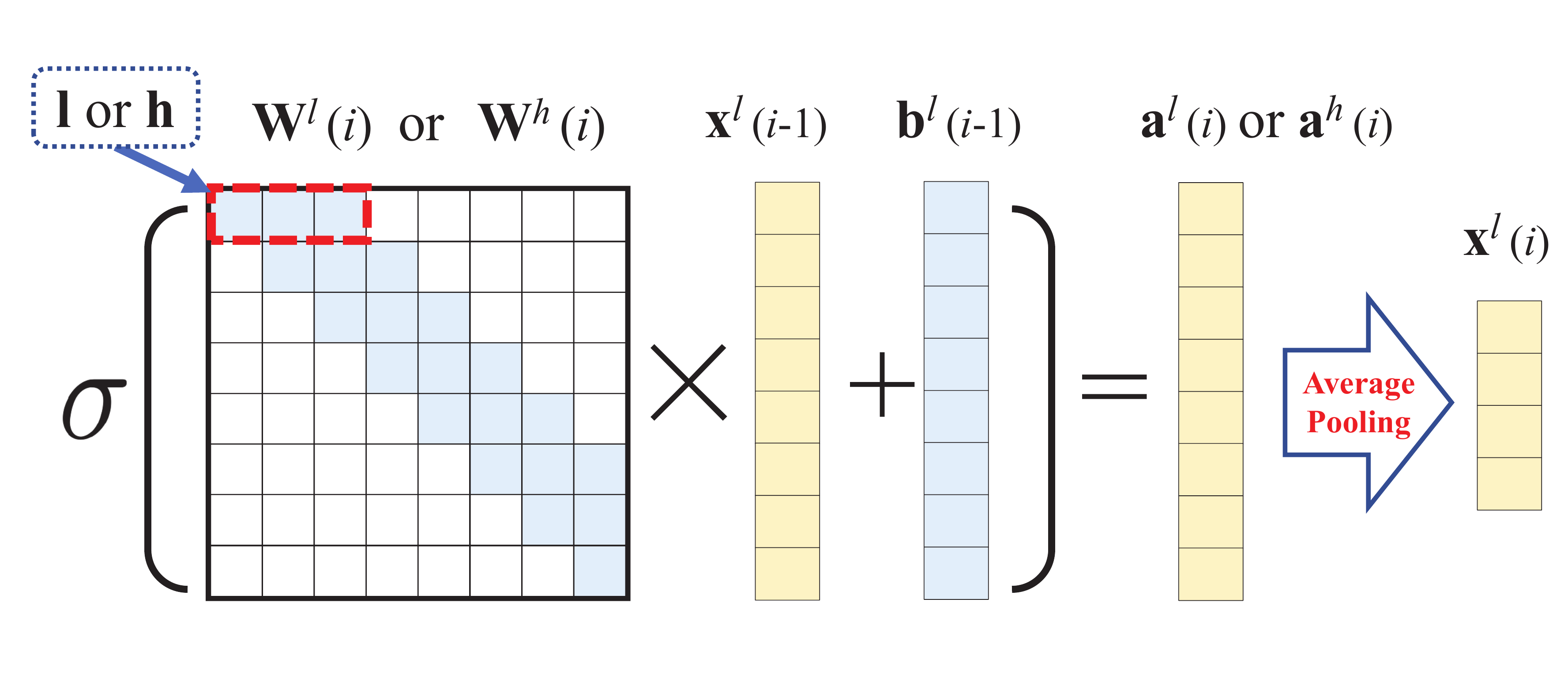}}
	\caption{The mWDN framework.}\label{fig:mWDN}
\end{figure}

\section{Model}

Throughout the paper, we use lowercase symbols such as $a$, $b$ to denote scalars, bold lowercase symbols such as $\mathbf{a}, \mathbf{b}$ to denote vectors, bold uppercase symbols such as $\mathbf{A}, \mathbf{B}$ to denote matrices, and uppercase symbols such as $A, B$, to denote constant.

\subsection{Multilevel Discrete Wavelet Decomposition}


Multilevel Discrete Wavelet Decomposition (MDWD)~\cite{mallat1989theory} is a wavelet based discrete signal analysis method, which can extract multilevel time-frequency features from a time series by decomposing the series as low and high frequency sub-series level by level.

We denote the input time series as $\mathbf{x} = \{x_1,$ $\ldots,$ $x_t,$ $\ldots, x_T \}$, and the low and high sub-series generated in the $i$-th level as $\mathbf{x}^{l}(i)$ and $\mathbf{x}^{h}(i)$. In the $(i+1)$-th level, MDWD uses a low pass filter $\mathbf{l}=\{l_1,$ $\ldots, l_k,$ $\ldots, l_K\}$ and a high pass filter $\mathbf{h}=\{h_1,$ $\ldots, h_k,$ $\ldots, h_K\}$, $K\ll T$, to convolute low frequency sub-series of the upper level as
\begin{equation}\label{eq:dwt}
\begin{aligned}
{a}_n^{l}(i+1) = \sum_{k=1}^K {x}_{n+k-1}^{l}(i)\cdot l_{k},\\
{a}_n^{h}(i+1) = \sum_{k=1}^K {x}_{n+k-1}^{l}(i)\cdot h_{k},
\end{aligned}
\end{equation}
where ${x}_n^{l}(i)$ is the $n$-th element of the low frequency sub-series in the $i$-th level, and $\mathbf{x}^{l}(0)$ is set as the input series. The low and high frequency sub-series $\mathbf{x}^{l}(i)$ and $\mathbf{x}^{h}(i)$ in the level $i$ are generated from the 1/2 down-sampling of the intermediate variable sequences $\mathbf{a}^{l}(i) = \left\{a^{l}_1(i), a^{l}_2(i), \ldots\right\}$ and $\mathbf{a}^{h}(i)= \left\{a^{h}_1(i), a^{h}_2(i), \ldots\right\}$.

The sub-series set $\mathcal{X}(i) = \left\{\mathbf{x}^{h}(1),\right.$ $\mathbf{x}^{h}(2),$ $\ldots,$ $\mathbf{x}^{h}(i),$  $\left. \mathbf{x}^{l}(i)\right\}$ is called as the $i$-th level decomposed results of $\mathbf{x}$. Specifically, $\mathcal{X}(i)$ satisfies: 1) We can fully reconstruct $\mathbf{x}$ from $\mathcal{X}(i)$; 2) The frequency from $\mathbf{x}^{h}(1)$ to $\mathbf{x}^{l}(i)$ is from high to low; 3) For different layers, $\mathcal{X}(i)$ has different time and frequency resolutions. As $i$ increases, the frequency resolution is increasing and the time resolution, especially for low frequency sub-series, is decreasing.

Because the sub-series with different frequencies in $\mathcal{X}$ keep the same order information with the original series $\mathbf{x}$, MDWD is regarded as time-frequency decomposition.

\subsection{Multilevel Wavelet Decomposition Network}

In this section, we propose a multilevel Wavelet Decomposition Network (mWDN), which approximatively implements a MDWD under a deep neural network framework.

The structure of mWDN is illustrated in Fig.~\ref{fig:mWDN}. As shown in the figures, the mWDN model hierarchically decomposes a time series using the following two functions
\begin{equation}\label{eq:wd_layer}
\begin{aligned}
&\mathbf{a}^l(i) = \sigma \left( \mathbf{W}^l(i) \mathbf{x}^l({i-1}) + \mathbf{b}^l(i) \right),\\
&\mathbf{a}^h(i) = \sigma \left( \mathbf{W}^h(i) \mathbf{x}^l({i-1}) + \mathbf{b}^h(i) \right),
\end{aligned}
\end{equation}
where $\sigma(\cdot)$ is a sigmoid activation function, and $\mathbf{b}^l(i)$ and $b^h(i)$ are trainable bias vectors initialized as close-to-zero random values. We can see the functions in Eq.~\eqref{eq:wd_layer} have similar forms as the functions in~Eq.~\eqref{eq:dwt} for MDWD. $\mathbf{x}^l(i)$ and $\mathbf{x}^h(i)$ also denote the low and high frequency sub-series of $\mathbf{x}$ generated in the $i$-th level, which are down-sampled from the intermediate variables $\mathbf{a}^{l}(i)$ and $\mathbf{a}^{h}(i)$ using
an average pooling layer as ${x}^l_j(i) = ({a}^l_{2j}(i) + {a}^l_{2j-1}(i))/2$.

In order to implement the convolution defined in Eq.~\eqref{eq:dwt}, we set the initial values of the weight matrices $\mathbf{W}^{l}$ and $\mathbf{W}^{h}$ as
\begin{equation}\label{eq:w_l}\centering
\begin{aligned}
\mathbf{W}^{l}(i) &  = \begin{bmatrix}
l_1&  l_2&  l_3&   \cdots & l_K&  \epsilon & \cdots &  \epsilon \\
\epsilon &  l_1&  l_2&    \cdots & l_{K-1} & l_K & \cdots & \epsilon  \\
\vdots &  \vdots &    \vdots &  \ddots &  \vdots &\vdots &  \vdots &\vdots  \\
\epsilon & \epsilon  & \epsilon &  \cdots &  l_1 & \cdots &  l_{K-1} & l_{K} \\
\vdots &  \vdots &    \vdots &  \ddots &  \vdots &\vdots &  \vdots &\vdots  \\
\epsilon & \epsilon  & \epsilon &  \cdots &  \cdots & \cdots &  l_{1} & l_{2} \\
\epsilon & \epsilon  & \epsilon &  \cdots &  \cdots & \cdots &  \epsilon & l_{1}
\end{bmatrix},
\end{aligned}
\end{equation}
\begin{equation}\label{eq:w_h}\centering
\begin{aligned}
\mathbf{W}^{h}(i) & = \begin{bmatrix}
h_1&  h_2&  h_3&   \cdots & h_K&  \epsilon & \cdots &  \epsilon \\
\epsilon &  h_1&  h_2&    \cdots & h_{K-1} & h_K & \cdots & \epsilon  \\
\vdots &  \vdots &    \vdots &  \ddots &  \vdots &\vdots &  \vdots &\vdots  \\
\epsilon & \epsilon  & \epsilon &  \cdots &  h_1 & \cdots &  h_{K-1} & h_{K} \\
\vdots &  \vdots &    \vdots &  \ddots &  \vdots &\vdots &  \vdots &\vdots  \\
\epsilon & \epsilon  & \epsilon &  \cdots &  \cdots & \cdots &  h_{1} & h_{2} \\
\epsilon & \epsilon  & \epsilon &  \cdots &  \cdots & \cdots &  \epsilon & h_{1}\end{bmatrix}.
\end{aligned}
\end{equation}
Obviously, $\mathbf{W}^{l}(i)$ and $\mathbf{W}^{h}(i)$ $\in \mathbb{R}^{P\times P}$, where $P$ is the size of $\mathbf{x}^l(i-1)$. The $\epsilon$ in the weight matrices are random values that satisfy $| \epsilon | \ll |l|, \forall l\in \mathbf{l}$ and $| \epsilon | \ll |h|, \forall h\in \mathbf{h}$. We use the Daubechies 4 Wavelet~\cite{rowe1995daubechies} in our practice, where the filter coefficients are set as
\begin{equation*}\label{eq:db4}\centering
\begin{aligned}
\mathbf{l} =& \{-0.0106,  0.0329,  0.0308, -0.187, \\
&-0.028 ,  0.6309,  0.7148, 0.2304\}, \\
\mathbf{h} =& \{-0.2304,  0.7148, -0.6309, -0.028, \\
&0.187 ,  0.0308, -0.0329, -0.0106\}.
\end{aligned}
\end{equation*}

From Eq.~\eqref{eq:wd_layer} to Eq.~\eqref{eq:w_l}, we use the deep neural network framework to implement an approximate MDWD. It is noteworthy that although the weight matrices $\mathbf{W}^{l}(i)$ and $\mathbf{W}^{h}(i)$ are initialized as the filter coefficients of MDWD, they are still trainable according to real data distributions.

\begin{figure}[t]
	\begin{center}
		\includegraphics[width=0.95\linewidth]{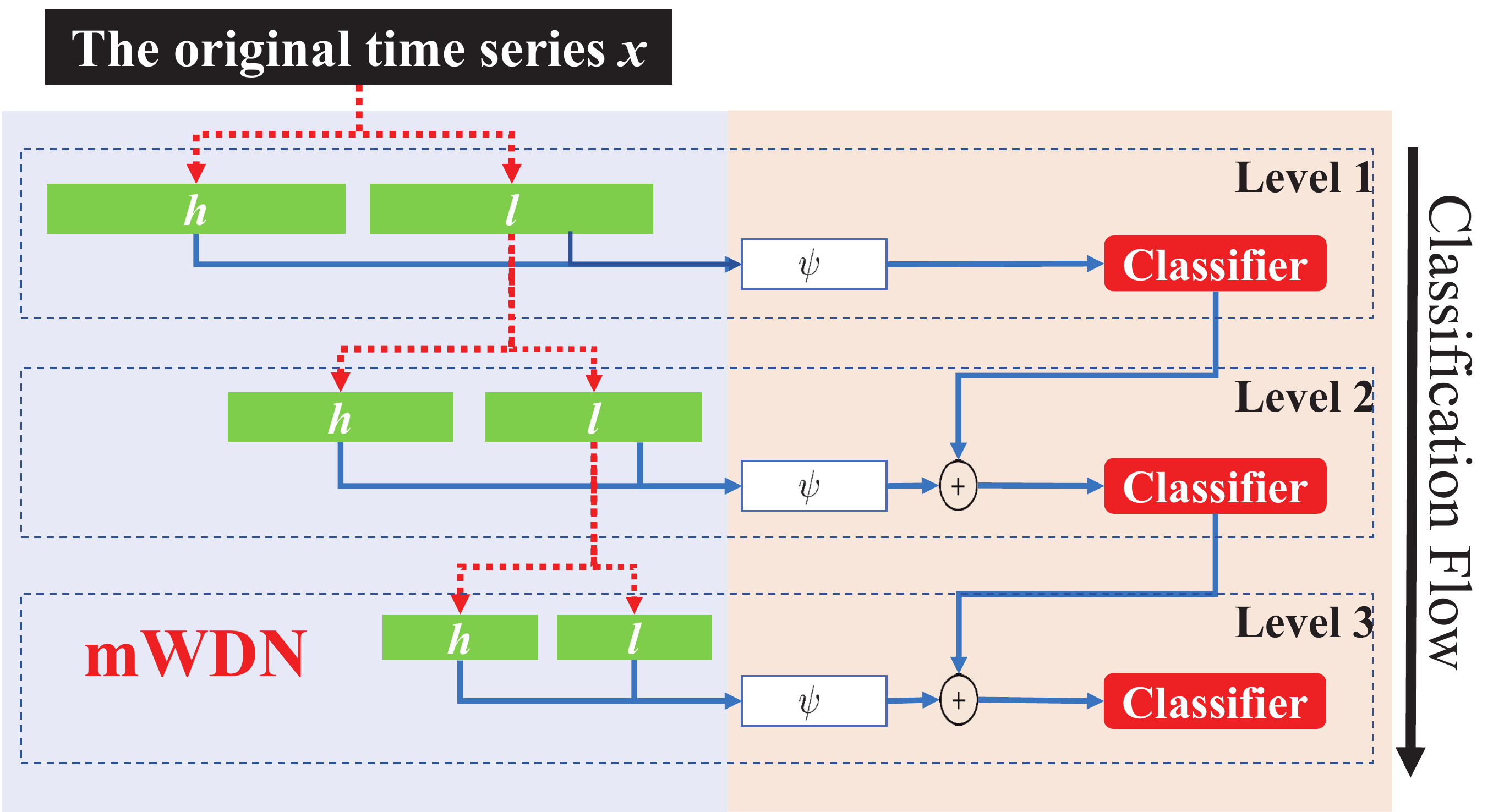}
	\end{center}
	\caption{The RCF framework.}
	\label{fig:framework}
\end{figure}

\subsection{Residual Classification Flow}

The task of TSC is to predict unknown category label of a time series. A key issue of TSC is extracting distinguishing features from time series data. The decomposed results $\mathcal{X}$ of mWDN are natural time-frequency features that could be used in TSC. In this subsection, we propose a Residual Classification Flow (RCF) network to exploit the potentials of mWDN in TSC.

The framework of RCF is illustrated in Fig.~\ref{fig:framework}. As shown in the figure, RCF
contains many independent classifiers. The RCF model connects the sub-series generated by the $i$-th mWDN level, {\it i.e.}, $\mathbf{x}^{h}(i)$ and $\mathbf{x}^{l}(i)$, with a forward neural network as
\begin{equation}\label{eq:basic_block}
\mathbf{u}(i) = \psi\left(\mathbf{x}^{h}(i),\mathbf{x}^{l}(i),\theta^{\psi}\right),
\end{equation}
where $\psi(\cdot)$ could be a multilayer perceptron, a convolutional network, or any other types of neural networks, and $\theta^{\psi}$ represents the trainable parameters. Moreover, RCF adopts a residual learning method~\cite{resnet} to join $\mathbf{u}(i)$ of all classifiers as
\begin{equation}\label{eq:resnet}
\mathbf{\hat{c}}({i}) = S\left(\mathbf{\hat{c}}({i-1}) + \mathbf{u}(i)\right),
\end{equation}
where $S(\cdot)$ is a softmax classifier, $\mathbf{\hat{c}}_{i}$ is a predicted value of one-hot encoding of the category label of the input series.


In the RCF model, the decomposed results of all mWDN levels, {\em i.e.} $\mathcal{X}(1), \ldots, \mathcal{X}(N)$, are evolved. Because the decomposed results in different mWDN levels have different time and frequency resolutions~\cite{mallat1989theory}, the RCF model can fully exploit patterns of the input time series from different time/frequency-resolutions. In other words, RCF employs a multi-view learning methodology to achieve high-performance time series classification.

Moreover, deep residual networks~\cite{resnet} were proposed to solve the problem that using deeper network structures may result in a great training difficulty. The RCF model also inherits this merit. In Eq.~\eqref{eq:resnet}, the $i$-th classifier makes decision based on $\mathbf{u}(i)$ and the decision made by the $(i-1)$-th classifier, which can learn from $\mathbf{u}(i)$ the incremental knowledge that the $(i-1)$-th classifier does not have. Therefore, users could append residual classifiers one after another until classification performance does not increase any more.

\begin{figure}\centering
	\includegraphics[width=0.95\linewidth]{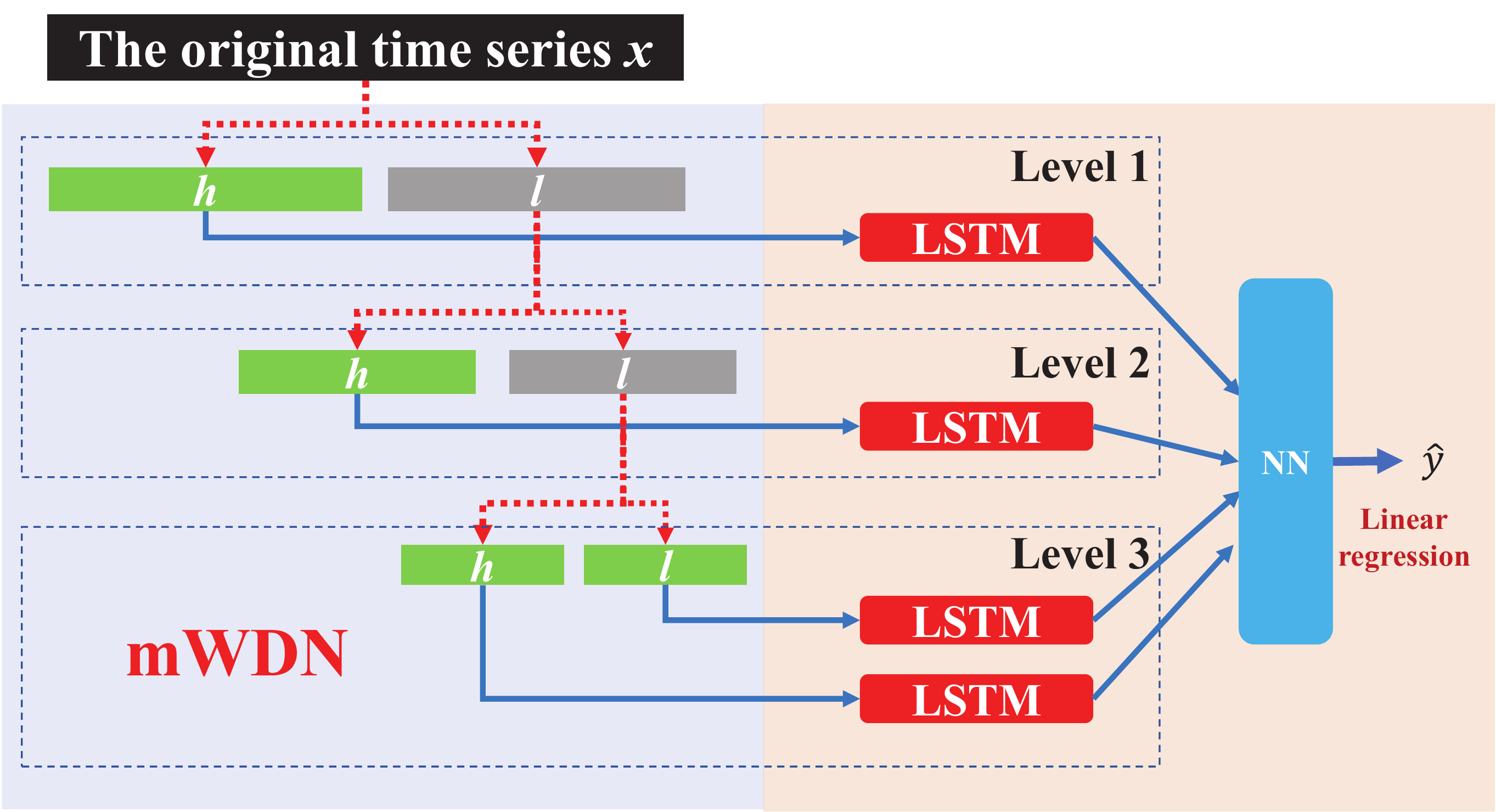}\\
	\caption{The mLSTM framework.}\label{fig:mlstm}
\end{figure}

\subsection{Multi-frequency Long Short-Term Memory}

In this subsection, we propose a multi-frequency Long-Short Term Memory (mLSTM) model based on mWDN for TSF. The design of mLSTM is based on the insight that the temporal correlations of points hidden in a time series have close relations with frequency. For example, large time scale correlations, such as long-term tendencies, usually lay in low frequency, and the small time scale correlations, such as short-term disturbances and events, usually lay in high frequency. Therefore, we could divide a complicated TSF problem as many sub-problems of forecasting sub-series decomposed by mWDN, which are relatively easier because the frequency components in the sub-series are simpler.

Given a time series with infinite length, on which we open a $T$ size slide window from the past to the time $t$ as
\begin{equation}\label{}
\mathbf{x} = \left\{x_{t-T+1}, \ldots, x_{t-1}, x_t\right\}.
\end{equation}
Using mWDN to decompose $\mathbf{x}$, we get the low and high frequency component series in the $i$-th level as
\begin{equation}\label{}
\begin{aligned}
&\mathbf{x}^l(i) = \{x^l_{t-\frac{T}{2^n}+1}(i), \ldots, x^l_{t-1}(i), x^l_t(i)\}, \\ &\mathbf{x}^h(i) = \{x^h_{t-\frac{T}{2^n}+1}(i), \ldots, x^h_{t-1}(i), x^h_t(i)\}.
\end{aligned}
\end{equation}
As shown in~Fig.~\ref{fig:mlstm}, the mLSTM model uses the decomposed results of the last level, {\it i.e.}, the sub-series in $\mathcal{X}(N) = \{\mathbf{x}^{h}(1),$ $\mathbf{x}^{h}(2),$ $\dots,$ $\mathbf{x}^{h}(N),$ $\mathbf{x}^{l}(N)\}$, as the inputs of $N+1$ independent LSTM sub-networks.
Every LSTM sub-network forecasts the future state of one sub-series in $\mathcal{X}(N)$. Finally,  a fully connected neural network is employed to fuse the LSTM sub-networks as an ensemble for forecasting.


\section{Optimization}

In TSC applications, we adopt a deep supervision method to train the RCF model~\cite{deep_supervision}. Given a set of time series $\{\mathbf{x}_1, \mathbf{x}_2, \ldots, \mathbf{x}_M\}$, we use cross-entropy as loss metric and define the objective function of the $i$-th classifier as
\begin{equation}\label{}
\mathcal{\tilde{J}}^{c}(i) = - \frac{1}{M} \sum_{m=1}^{M} \left( \mathbf{c}_m^\top \ln \hat{\mathbf{c}}_m(i) + (1-\mathbf{c}_m)^\top \ln(1-\hat{\mathbf{c}}_m(i)) \right),
\end{equation}
where $\mathbf{c}_m$ is the one-hot encoding of $\mathbf{x}_m$'s real category, and $\mathbf{\hat{c}}_m(i)$ is the softmax output of the $i$-th classifier with the input $\mathbf{x}_m$. For a RCF with $N$ classifiers, the final objective function is a weighted sum of all $\tilde{\mathcal{J}}(i)$~\cite{deep_supervision}:
\begin{equation}\label{eq:J_rcf}
\mathcal{J}^c = \sum_{i=1}^N \frac{i}{N} \mathcal{\tilde{J}}^c(i).
\end{equation}
The result of the last classifier, $\hat{\mathbf{c}}(N)$, is used as the final classification result of RCF.

In TSF applications, we adopt a pre-training and fine turning method to train the mLSTM model. In the pre-training step, we use MDWD to decompose the real value of the future state to be predicted as $N$ wavelet components, {\em i.e.} ${\mathbf{y}^p} = \{ {y}^h(1),$ ${y}^h(2),$ $\ldots,$ ${y}^h(N),$ ${y}^l(N)\}$, and then combine the outputs of all LSTM sub-network as $\hat{\mathbf{y}}^p$, then the objective function of the pre-training step is defined as
\begin{equation}\label{}
\mathcal{\tilde{J}}^f = - \frac{1}{M} \sum_{m=1}^M \| \mathbf{y}_m - \hat{\mathbf{y}}_m^p \|_F^2,
\end{equation}
where $\|\cdot\|_F$ is the Frobenius Norm. In the fine-turning step, we use the following objective function to train mLSTM based on the parameters learned in the pre-training step:
\begin{equation}\label{eq:J_mlstm}
\mathcal{{J}}^f = \frac{1}{T} \sum_{t=1}^{T} \left( \hat{y} - {y} \right)^2,
\end{equation}
where $\hat{y}$ is future state predicted by mLSTM and $y$ is the real value.

We use the error back propagation (BP) algorithm to optimize the objective functions. Denoting $\theta$ as the parameters of the RCF or mLSTM model, the BP algorithm iteratively updates $\theta$ as
\begin{equation}\label{eq:bp}
\theta \leftarrow \theta- \eta\frac{\partial \mathcal{J}(\theta)}{\partial \theta},
\end{equation}
where $\eta$ is an adjustable learning rate. The weight matrices $\mathbf{W}^h(i)$ and $\mathbf{W}^l(i)$  of mWDN are also trainable in Eq.~\eqref{eq:bp}. A problem of training parameters with preset initial values like $\mathbf{W}^l(i)$ and $\mathbf{W}^h(i)$ is that the model may ``forget'' the initial values~\cite{french1999catastrophic} in the training process. To deal with this, we introduce two regularization items to the objective function and therefore have
\begin{equation}\label{eq:obj}
\begin{aligned}
\mathcal{J}^* =   \mathcal{{J}}(\theta) & + \alpha \sum_i \| \mathbf{W}^l(i) - \tilde{\mathbf{W}}^l(i) \|^2_F \\
& \left. + \beta \sum_i \| \mathbf{W}^h(i) - \tilde{\mathbf{W}}^h(i) \|^2_F \right. ,
\end{aligned}
\end{equation}
where $\tilde{\mathbf{W}}^l(i)$ and $\tilde{\mathbf{W}}^h(i)$ are the same matrices as $\mathbf{W}^h(i)$ and $\mathbf{W}^h(i)$ except that $\epsilon = 0$, and $\alpha, \beta$ are hyper-parameters which are set as empirical values. Accordingly, the BP algorithm iteratively updates the weight matrices of mWDN as
\begin{equation}\label{}
\begin{aligned}
&\mathbf{W}^l(i) \leftarrow \mathbf{W}^l(i) - \eta \left( \frac{\partial \mathcal{J}}{\partial \mathbf{W}^l(i)} - 2\alpha \left( \mathbf{W}^l(i)- \mathbf{\tilde{W}}(i) \right)\right), \\
&\mathbf{W}^h(i) \leftarrow \mathbf{W}^h(i) - \eta \left( \frac{\partial \mathcal{J}}{\partial \mathbf{W}^h(i)} - 2\beta \left( \mathbf{W}^l(i)- \mathbf{\tilde{W}}(i) \right)\right).
\end{aligned}
\end{equation}
In this way, the weights in mWDN will converge to a point that is near to the wavelet decomposed prior, unless wavelet decomposition is far inappropriate to the task.

\begin{table*}[t!]\caption{Comparison of Classification Performance on 40 UCR Time Series Datasets}
	\label{tab:class_performance}	\centering   \small
	\begin{tabular}{r|ccccc|ccc|c}
		\toprule
		Err Rate & RNN  & LSTM  & MLP  & FCN  & ResNet  &  {\bf MLP-RCF} &  {\bf FCN-RCF} & {\bf ResNet-RCF} & Wavelet-RCF \\\midrule
		Adiac   &0.233&0.341& {0.248} & {\bred 0.143} &0.174&{0.212}& { 0.155}  & {\snd 0.151} &0.162\\
		Beef    &0.233&0.333&0.167&0.25&0.233& {\snd 0.06} & {\bred 0.03} & {\snd 0.06} &{\snd 0.06}\\
		CBF     &0.189&0.118&0.14& {\bred 0} &{\snd 0.006}& { 0.056} & {\bred 0}  & {\bred 0} &0.016\\
		ChlorineConcentration &0.135&0.16&0.128&0.157&0.172& { 0.096} & {\bred 0.068} & {\snd 0.07} &0.147\\
		CinCECGtorso &0.333&0.092&0.158&0.187&0.229& { 0.117} & {\snd 0.014} & { 0.084} &{\bred0.011}\\
		CricketX &0.449&0.382&0.431& {\snd 0.185} & {\bred 0.179} & { 0.321} &0.216&0.297&{  0.211}\\
		CricketY &0.415&0.318&0.405& {0.208} & { 0.195} & { 0.254} &{\bred 0.172}&0.301 &{\snd 0.192}\\
		CricketZ &0.4&0.328& {0.408} & {\snd 0.187} & { 0.187} & { 0.313}  &{\bred 0.162}& {0.275}&{\bred 0.162}\\
		DiatomSizeReduction &0.056&0.101&0.036&0.07&0.069& {\bred 0.013 }& {\snd 0.023}  & { 0.026} &0.028\\
		ECGFiveDays &0.088&0.417&0.03&{\snd 0.015}&0.045& { 0.023}  &  {\bred 0.01 }&  { 0.035}&0.016\\
		FaceAll &0.247&0.192&0.115& {\bred 0.071} &0.166& { 0.094} &0.098& { 0.126} &{\snd 0.076}\\
		FaceFour &0.102&0.364&0.17&{0.068}&0.068& { 0.102}  & {\bred 0.05}  & {\snd 0.057} &0.058\\
		FacesUCR &0.204&0.091&0.185& {\snd 0.052} & {\bred 0.042} & { 0.15} &0.087&0.102&{ 0.087}\\
		50words &0.316&0.284& { 0.288} &0.321&{\snd 0.273}&0.316& { 0.288}  &{\bred  0.258 }&0.3\\
		FISH &0.126&0.103&0.126& {0.029} & {\bred 0.011} & { 0.086}  &{\snd 0.021} &0.034&0.026\\
		GunPoint &0.1&0.147&0.067&{\bred 0 }& {\snd 0.007} & { 0.033} & {\bred 0 }&0.02&{\bred 0}\\
		Haptics &0.594&0.529&0.539& {\bred 0.449} &0.495& { 0.480}  &{\snd 0.461}& { 0.473} &0.476\\
		InlineSkate &0.667&0.638&0.649&0.589&0.635& {\bred 0.543} & {\snd 0.566}  & { 0.578} &0.572\\
		ItalyPowerDemand &0.055&0.072&0.034&0.03&0.04&  { 0.031} & {\bred0.023} & { 0.034} &{\snd 0.028}\\
		Lighting2 & {\bred 0} & {\bred 0}  &0.279& {0.197} &0.246& { 0.213} &{\snd 0.145}& { 0.197} &0.162\\
		Lighting7 &0.288&0.384&0.356&{\snd 0.137}& { 0.164} & { 0.179} & {\bred 0.091} &0.177&0.144\\
		MALLAT &0.119&0.127&0.064&{\bred  0.02} & {\snd 0.021} & { 0.058} &0.044&0.046&{ 0.024}\\
		MedicalImages &0.299&0.276&0.271&0.208&0.228& { 0.251} & {\bred 0.164 } & {\snd 0.188} &0.206\\
		MoteStrain &0.133&0.167&0.131&{\snd 0.05}&0.105& { 0.105}  &  {0.076} & {\bred 0.032} &{ 0.05}\\
		NonInvasiveFatalECGThorax1 &0.09&0.08&0.058&0.039&0.052& {\snd 0.029} & {\bred  0.026} & { 0.04} &0.042\\
		NonInvasiveFatalECGThorax2 &0.069&0.071&0.057&0.045&0.049& { 0.056} & {\bred 0.028 }& {\snd 0.033} &0.048\\
		OliveOil &0.233&0.267&0.6&0.167&0.133& { 0.03} & {\bred  0} &{\bred  0 }&{\snd 0.012}\\
		OSULeaf &0.463&0.401&0.43& {\bred  0.012} &{0.021}& { 0.342} &  {\snd 0.018} &{ 0.021}& { 0.021}\\
		SonyAIBORobotSurface &0.21&0.309& { 0.273} &{\snd 0.032}& {\bred 0.015} & { 0.193}  &0.042&0.032&0.052\\
		SonyAIBORobotSurfaceII &0.219&0.187&0.161& {\bred 0.038} & {\bred 0.038} & { 0.092}  &{\snd 0.064}&0.083&0.072\\
		StarLightCurves &0.027&0.035&0.043&0.033&0.029& {\snd 0.021}  &{\bred 0.018}& { 0.027} &0.03\\
		SwedishLeaf &0.085&0.128&0.107& {\snd 0.034} &0.042& { 0.089} &0.057& {\bred 0.017} &{ 0.046}\\
		Symbols &0.179&0.117&0.147& {\bred  0.038 }&0.128& { 0.126} & {\snd 0.04} & { 0.107}  &{0.084}\\
		TwoPatterns &0.005&{\snd 0.001}&0.114&0.103& {\bred 0} & { 0.070}  & {\bred 0}  & {\bred 0} &0.005\\
		uWaveGestureLibraryX &0.224&0.195&0.232&0.246&0.213& { 0.213} &  { 0.218} & {\snd 0.194 }&{\bred 0.162}\\
		uWaveGestureLibraryY &0.335& { 0.265} & { 0.297} & {0.275} &0.332&0.306& {\bred 0.232}& { 0.296} &{\snd 0.241}\\
		uWaveGestureLibraryZ &0.297&0.259& { 0.295} &0.271&0.245&0.298& { 0.265} &  {\snd 0.204 } &{\bred 0.194}\\
		wafer & {\bred 0} & {\bred 0} &0.004&{\snd 0.003}& {\snd 0.003}& {\snd 0.003}  & {\bred  0 }& {\bred  0 }&{\bred 0}\\
		WordsSynonyms &0.429&0.343&0.406&0.42& { 0.368} & { 0.391}  &{\snd 0.338 } &0.387&{\bred 0.314}\\
		yoga &0.202&0.158&0.145&0.155&0.142& { 0.138}  & {\bred  0.112  }& { 0.139} &{\snd 0.128}\\\midrule
		Winning times        & 2  & 2  & 0  & {\snd 9}  & 6  & 2  & {\bf 19}  & 7 & 7\\
		AVG arithmetic ranking & 7.425	& 6.825	& 7.2	& 4.025	& 4.55	& 5.15	& {\bf 2.175}	& 3.375	 & {\snd 3.075}\\
		AVG geometric ranking  & 6.860	& 6.131	& 7.043	& 3.101	& 3.818	& 4.675	& {\bf 1.789}	& 2.868	 & {\snd 2.688}\\
		MPCE  &0.039   &0.043   &0.041   &0.023   &0.025   &0.028   &{\bf 0.017}   &0.021 &{\snd 0.019}  \\
		\bottomrule
	\end{tabular}
\end{table*}


\section{Experiments}

In this section, we evaluate the performance of the mWDN-based models in both the TSC and TSF tasks. 

\subsection{Task I: Time Series Classification}
\label{subsec:Classification}

{\bf Experimental Setup.}~The classification performance was tested on 40 datasets of the UCR time series repository~\cite{UCRArchive}, with various competitors as follows:

\begin{itemize}
	\item {\it RNN} and {\it LSTM}. Recurrent Neural Networks~\cite{rnn}, and Long Short-Term Memory~\cite{lstm} are two kinds of classical deep neural networks widely used in time series analysis.
	\item {\it MLP}, {\it FCN}, and {\it ResNet}. These three models were proposed in~\cite{baselines} as strong baselines on the UCR time series datasets. They have the same framework: an input layer, followed by three hidden basic blocks, and finally a softmax output. MLP adopts a fully-connected layer as its basic block, FCN and ResNet adopt a fully convolutional layer and a residual convolutional network, respectively, as their basic blocks.
	\item{\it MLP-RCF}, {\it FCN-RCF}, and {\it ResNet-RCF}. The three models use the basic blocks of MLP/FCN/ResNet as the $\psi$ model of RCF in Eq.~\eqref{eq:basic_block}. We compare them with MPL/FCN/ResNet to verify the effectiveness of RCF.
	\item{\it Wavelet-RCF}. This model has the same structure as ResNet-RCF but replaces the mWDN part with a standard MDWD with fixed parameters. We compare it with ResNet-RCF to verify the effectiveness of trainable parameters in mWDM.
\end{itemize}

For each dataset, we ran a model 10 times and returned the average {\it classification error rate} as the evaluation. To compare the overall performances on all the 40 data sets, we further introduced {\it Mean Per-Class Error} (MPCE) as the performance indicator for each competitor~\cite{baselines}. Let $C_k$ denote the amount of categories in the $k$th dataset, and $e_k$ the error rate of a model on that dataset, MPCE of a model is then defined as
\begin{equation}\label{}
\mathrm{MPCE} = \frac{1}{K} \sum_{l=1}^K \frac{e_k}{C_k}.
\end{equation}
Note that the factor of category amount is wiped out in MPCE. A smaller MPCE value indicates a better overall performance.

\noindent {\bf Results \& Analysis.}~Table~\ref{tab:class_performance} shows the experimental results, with the summarized information listed in the bottom two lines. Note that the best performance for each dataset is highlighted in bold, and the second best is in italic. From the table, we have various interesting observations. Firstly, it is clear that among all the competitors, FCN-RCF achieves the best performance in terms of both the largest number of wins (the best in 19 out of 40 datasets) and the smallest MPCE value. While the baseline FCN itself also achieves a satisfactory performance --- the second largest number of wins at 9 and a rather small MPCE value at 0.023, the gap to FCM-RCF is still rather big, implying the significant benefit from adopting our RCF framework. This is actually not an individual case; from Table~\ref{tab:class_performance}, MLP-RCF performs much better than MLP on 37 datasets, and the number for ResNet-RCF against ResNet is 27. This indicates RCF is indeed a general framework compatible with different types of deep learning classifiers and can improve TSF performance sharply.

Another observation is from the comparison between Wavelet-RCF and ResNet-RCF. Table~\ref{tab:class_performance} shows that Wavelet-RCF achieved the second overall performance on MPCE and AVG rankings, which indicates that the frequency information introduced by wavelet tools is very helpful for time series problems. It is clear from the table that ResNet-RCF outperforms Wavelet-RCF on most of the datasets. This strongly demonstrates the advantage of our RCF framework in adopting parameter-trainable mWDN under the deep learning architecture, rather than using directly the wavelet decomposition as a feature engineering tool. More technically speaking, compared with Wavelet-RCF, mWND-based ResNet-RCF can achieve a good tradeoff between the prior of frequency-domain and the likelihoods of training data. This well illustrates why RCF based models can achieve much better results in the previous observation.

{\bf Summary.}~The above experiments demonstrate the superiority of RCF based models to some state-of-the-art baselines in the TSC tasks. The experiments also imply that the trainable parameters in a deep learning architecture and the strong priors from wavelet decomposition are two key factors for the success of RCF. 

\begin{figure}[t]\centering
	\begin{center}\centering
		\subfigure[Comparison by MAPE]{\label{fig:2nd_iLvar_MAE} \includegraphics[width=0.23\textwidth]{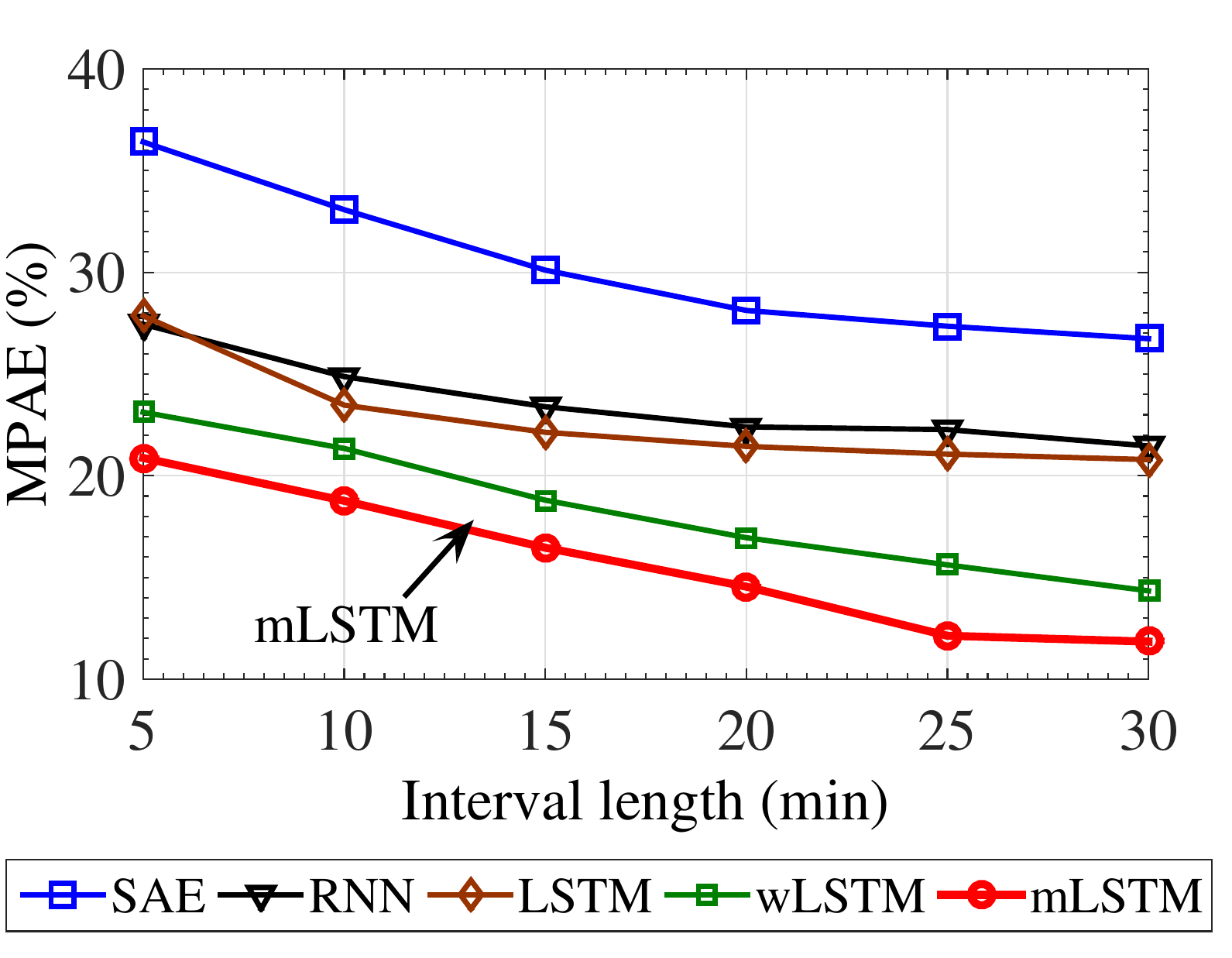}}
		\subfigure[Comparison by RMSE]{\label{fig:2nd_iLvar_MAPE}
			\includegraphics[width=0.23\textwidth]{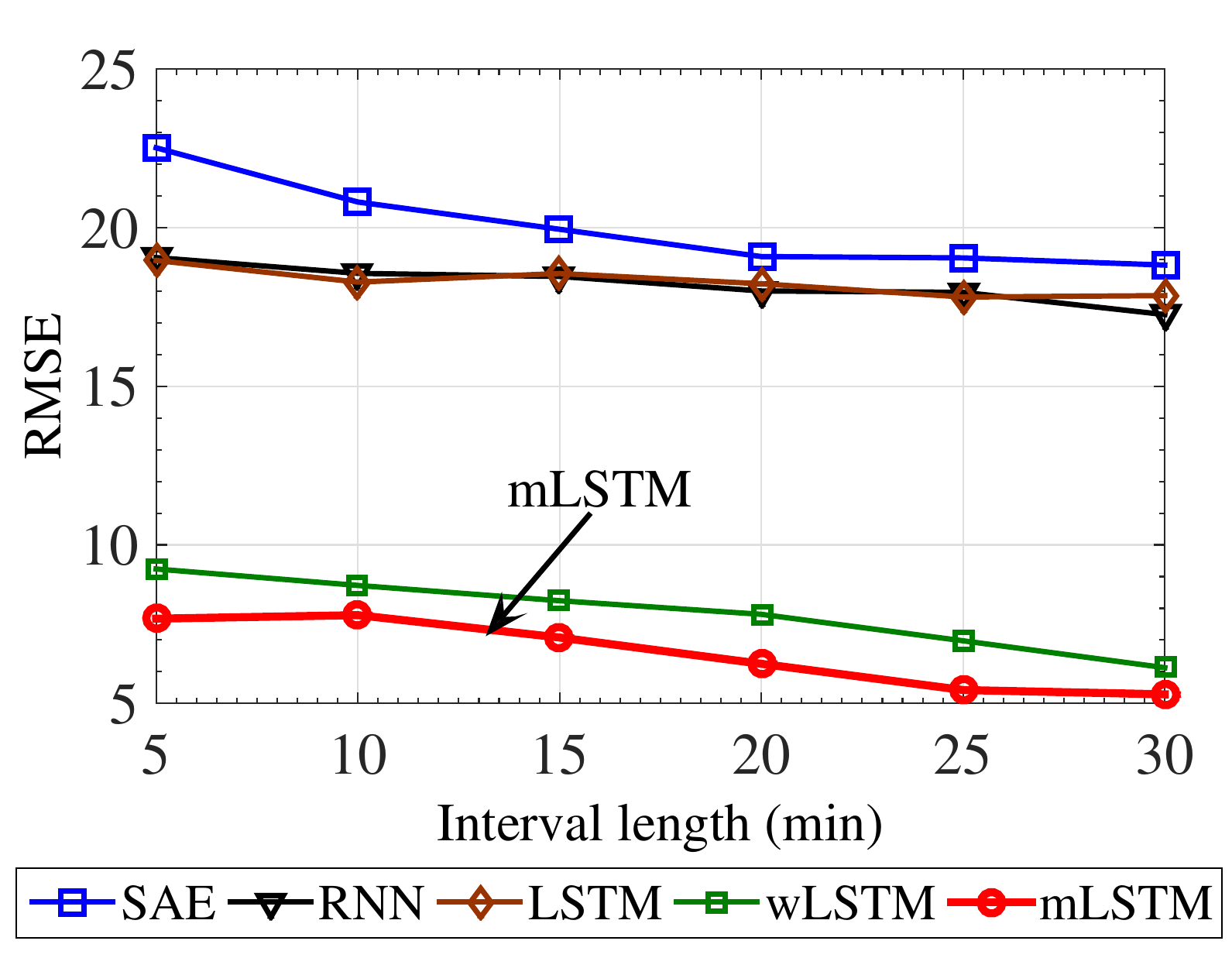}}
	\end{center}
	\caption{Comparison of prediction performance with varying period lengths (Scenario I).}
	\label{fig:period}
\end{figure}

\begin{figure}[t]\centering
	\begin{center}\centering
		\subfigure[Comparison by MAPE]{\label{fig:2nd_sLvar_MAE}
			\includegraphics[width=0.23\textwidth]{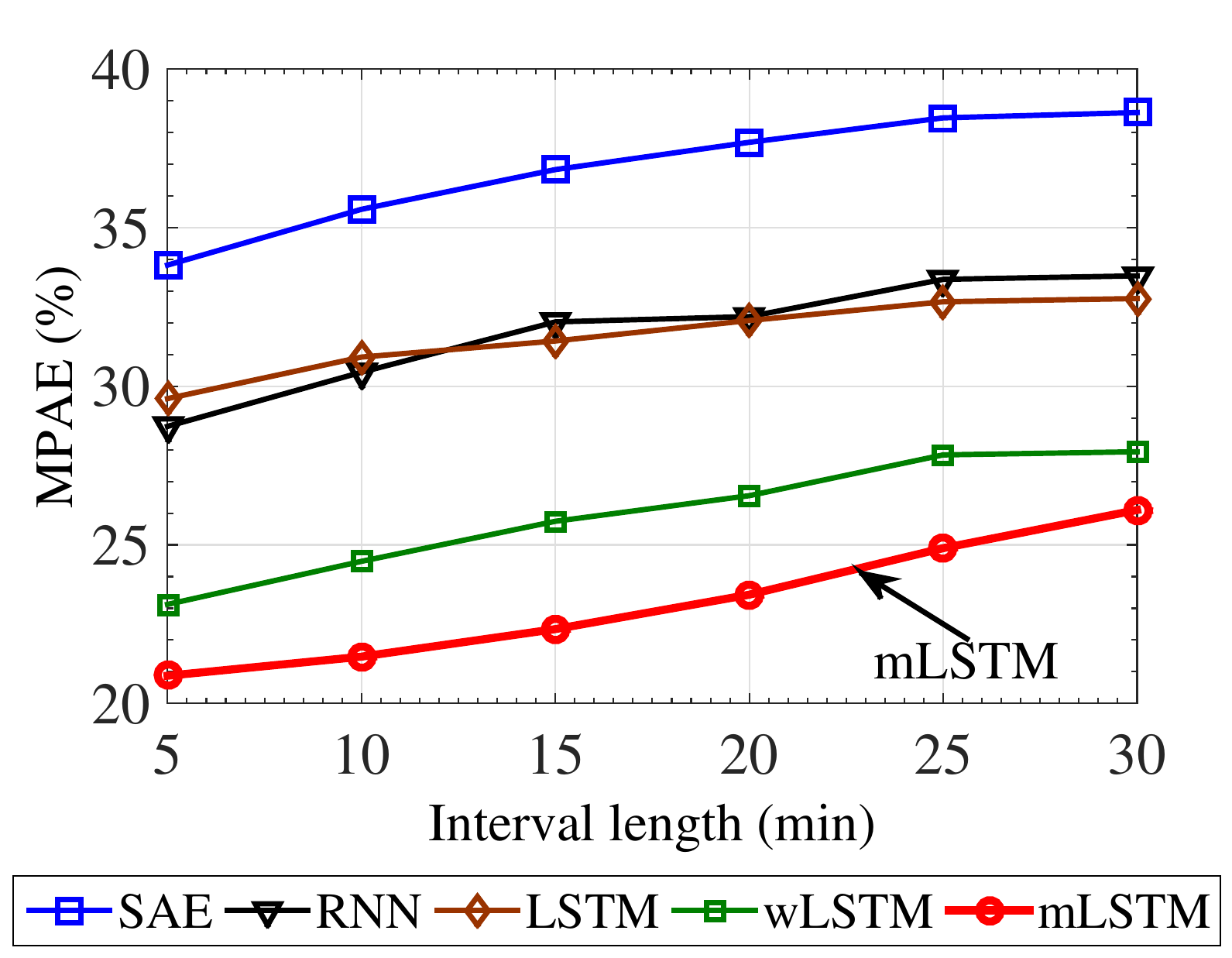}}
		\subfigure[Comparison by RMSE]{\label{fig:2nd_sLvar_MAPE}
			\includegraphics[width=0.23\textwidth]{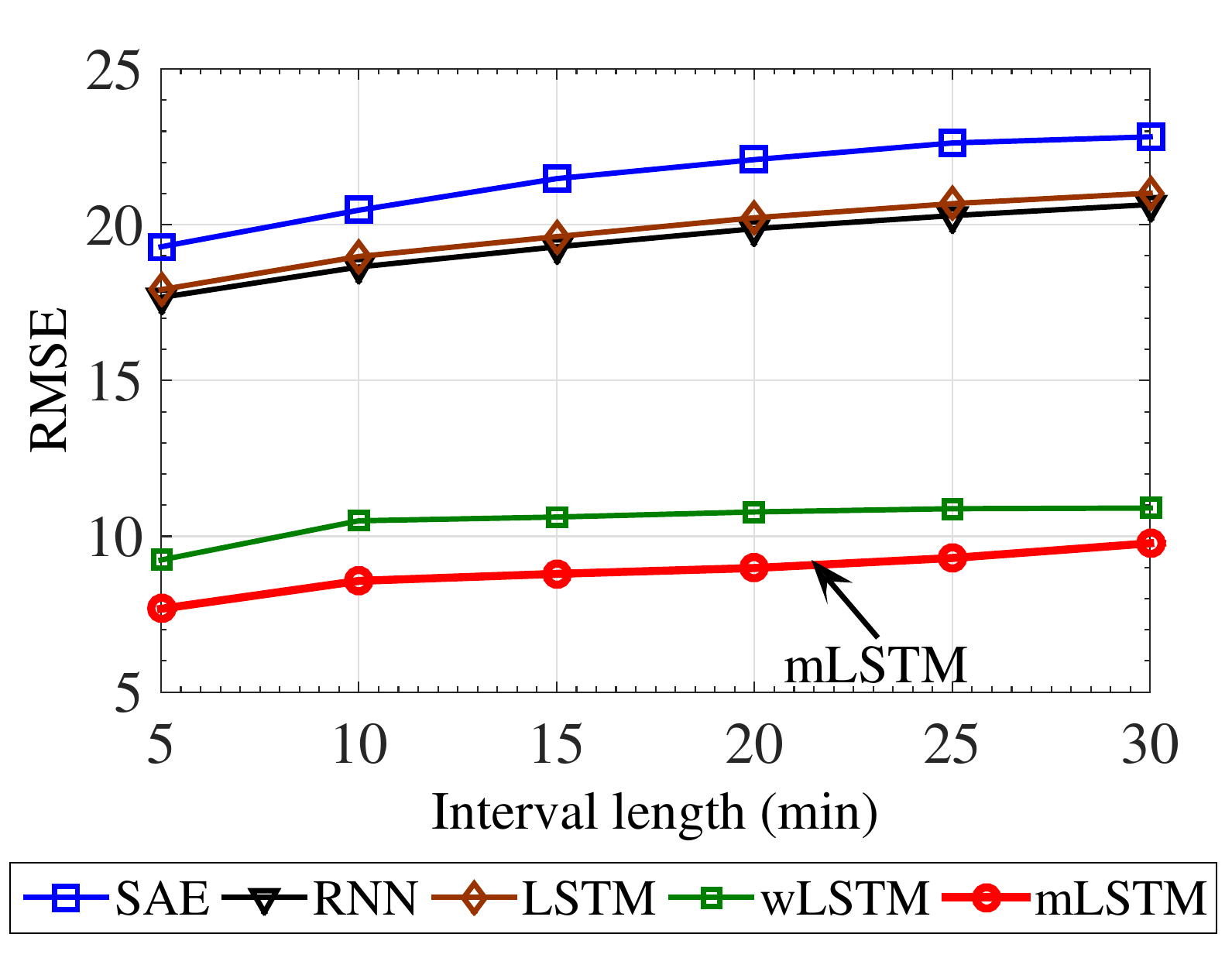}}
	\end{center}
	\caption{Comparison of prediction performance with varying interval lengths (Scenario II).}
	\label{fig:interval}
\end{figure}

\subsection{Task II: Time Series Forecasting}
\label{subsec:Forecasting}

{\bf Experimental Setup.}~We tested the predictive power of mLSTM on a visitor volume prediction scenario~\cite{wang2}. The experiment adopts a real-life dataset named \textit{WuxiCellPhone}, which contains user-volume time series of 20 cell-phone base stations located in the downtown of Wuxi city during two weeks. Detail informantion of cell-phone data refers~\cite{wang3,wang1,Song2017Recovering}.  The time granularity of a user-volume series is 5 minutes.
In the experiments, we compared mLSTM with the following baselines:
\begin{itemize}
	\item {\it SAE} (Stacked Auto-Encoders), which has been used in various TSF tasks~\cite{lv2015traffic}.
	\item {\it RNN} (Recurrent Neural Networks) and {\it LSTM} (Long Short-Term Memory), which are specifically designed for time series analysis.
	\item {\it wLSTM} , which has the same structure with mLSTM but replaces the mWDN part with a standard MDWD.
\end{itemize}

We use three metrics to evaluate the performance of the models, including {\it Mean Absolute Percentage Error} (MAPE) and {\it Root Mean Square Error} (RMSE), which are defined as
\begin{equation}
\begin{aligned}
\mathrm{MAPE} &= \frac{1}{T} \sum_{t=1}^T \frac{\left|  \hat{x}_t - x_t \right|}{x_t}\times 100\%, \\
\mathrm{RMSE} &= \sqrt{\frac{1}{T}\sum_{t=1}^T\left(\hat{x}_t-x_t\right)^2},
\end{aligned}
\end{equation}
where $x_t$ is the real value of the $t$-th sample in a time series, and $\hat{x}_t$ is the predicted one. The less value of the three metrics means the better performance.

{\bf Results \& Analysis.}~We compared the performance of the competitors in two TSF scenarios suggested in~\cite{icdm}. In the first scenario, we predicted the average user volumes of a base station in subsequent periods. The length of the periods was varied from 5 to 30 minutes. Fig.~\ref{fig:period} is a comparison of the performance averaged on the 20 base stations in one week. As can be seen, while all the models experience a gradual decrease in prediction error as the period length increases, that mLSTM achieves the best performance compared with the baselines. Particularly, the performance of mLSTM is consistently better than wLSTM, which again approves the introduction of mWDN for time series forecasting.

In the second scenario, we predicted the average user volumes in 5 minutes after a given time interval varying from 0 to 30 minutes. Fig.~\ref{fig:interval} is a performance comparison between mLSTM and the baselines. Different from the tend we observed in Scenario I, the prediction errors in Fig.~\ref{fig:interval} generally increase along the x-axis for the increasing uncertainty. From Fig.~\ref{fig:interval} we can see that mLSTM again outperforms wLSTM and other baselines, which confirms the observations from Scenario I.

{\bf Summary.}~The above experiments demonstrate the superiority of mLSTM to the baselines. The mWDN structure adopted by mLSTM again becomes an important factor for the success.

\begin{figure}[t]\centering
	\begin{center}\centering
		\subfigure[Cell-phone User Number]{\label{fig:wuxi_org}
			\includegraphics[width=0.22\textwidth]{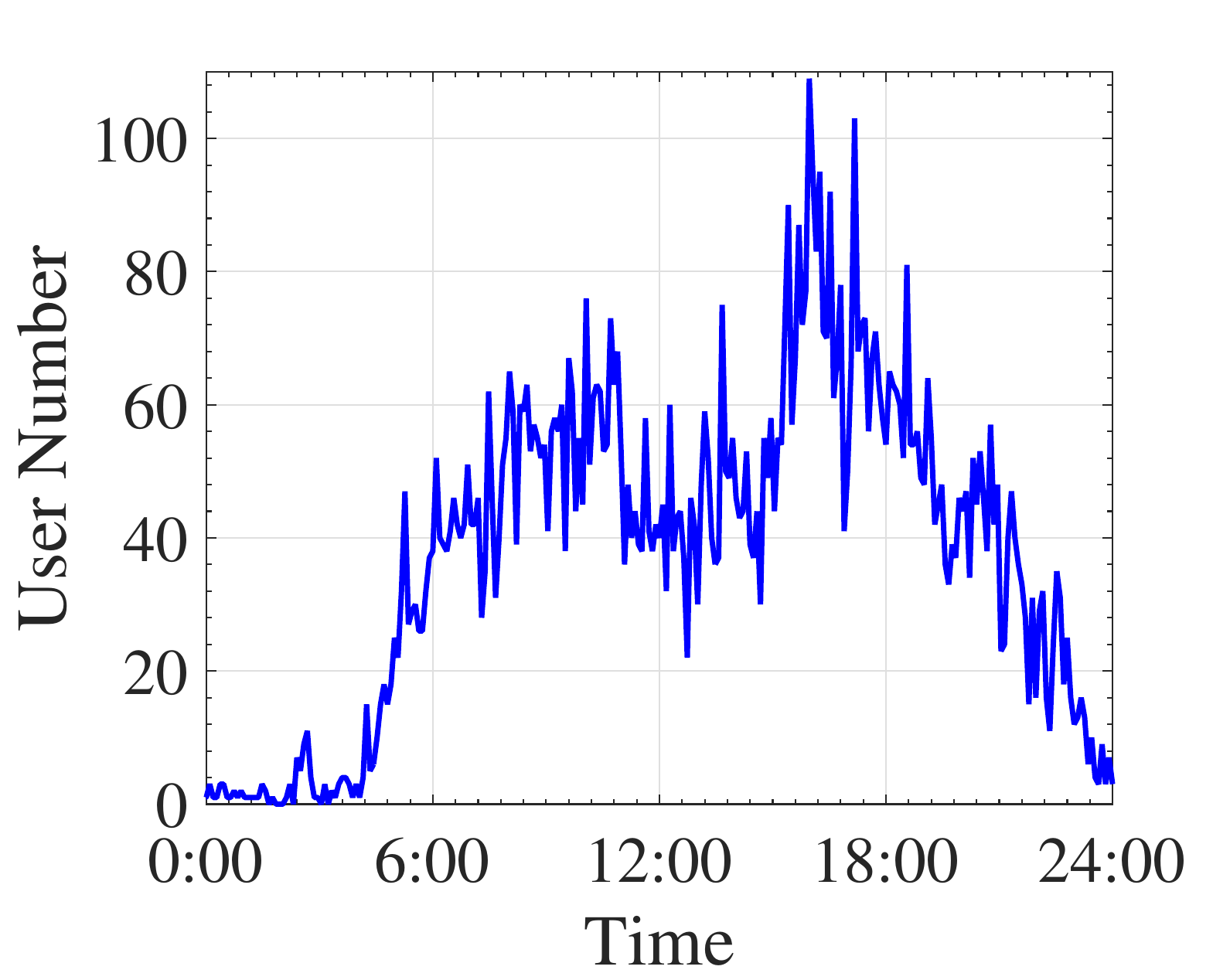}}
		\subfigure[ECG]{\label{fig:ecg_org}
			\includegraphics[width=0.22\textwidth]{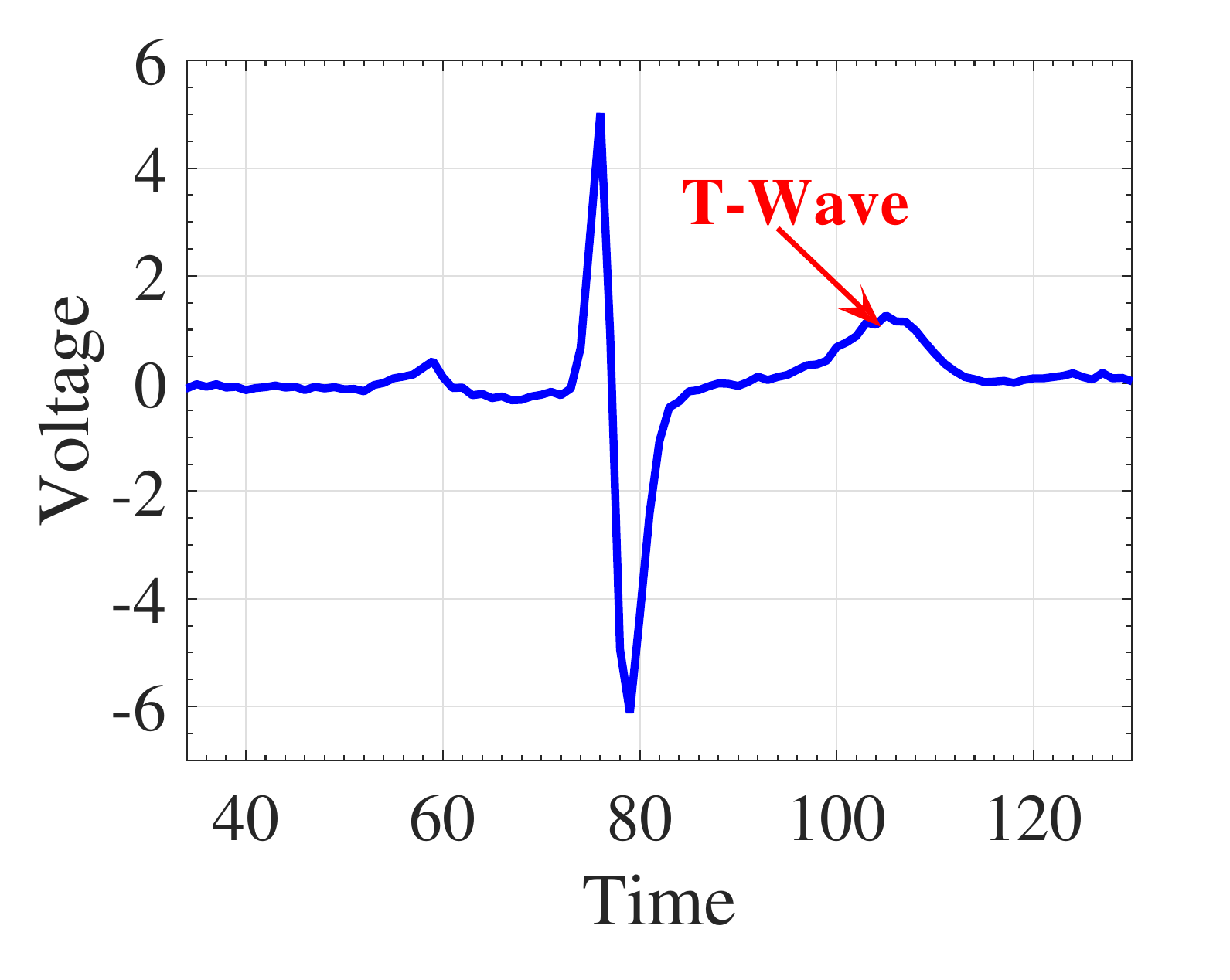}}
	\end{center}
	\vspace{-0.4cm}
	\caption{Samples of time series.}
	\label{fig:samples}
\end{figure}

\section{Interpretation}
In this section, we highlight the unique advantage of our mWDN model: the interpretability. Since mWDN  is embedded with a discrete wavelet decomposition, the outputs of the middle layers in mWDN, {\it i.e.}, $\mathbf{x}^l(i)$ and $\mathbf{x}^h(i)$, inherit the physical meanings of wavelet decompositions. We here take two data sets for illustration: \textit{WuxiCellPhone} used in Sect.~4.2 and \textit{ECGFiveDays} used in Sect.~4.1. Fig.~\ref{fig:wuxi_org} shows a sample of the user number series of a cell-phone base station in one day, and Fig.~\ref{fig:ecg_org} exhibits an electrocardiogram (ECG) sample.

\subsection{The Motivation}
\begin{figure*}[t!]
	\centering
	\subfigure[Cell-phone User Numbers in Different Layers]{\label{fig:wuxi_wavelet}
		\includegraphics[width=0.95\columnwidth]{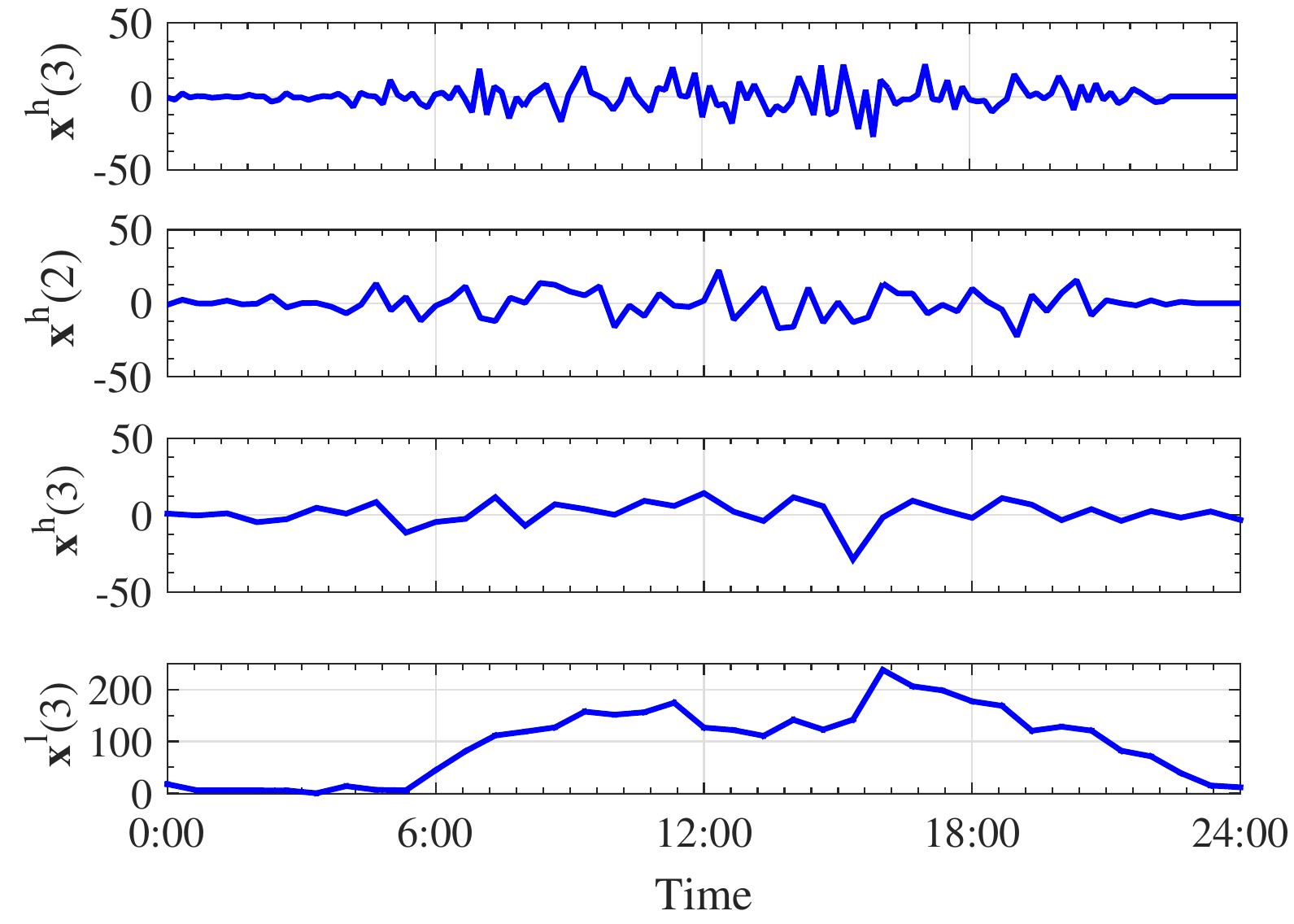}}\;\;\;\;\;\;\;\;\;\;\;\;\;\;
	\subfigure[ECG Waves in Different Layers]{\label{fig:ecg_wavelet}
		\includegraphics[width=0.95\columnwidth]{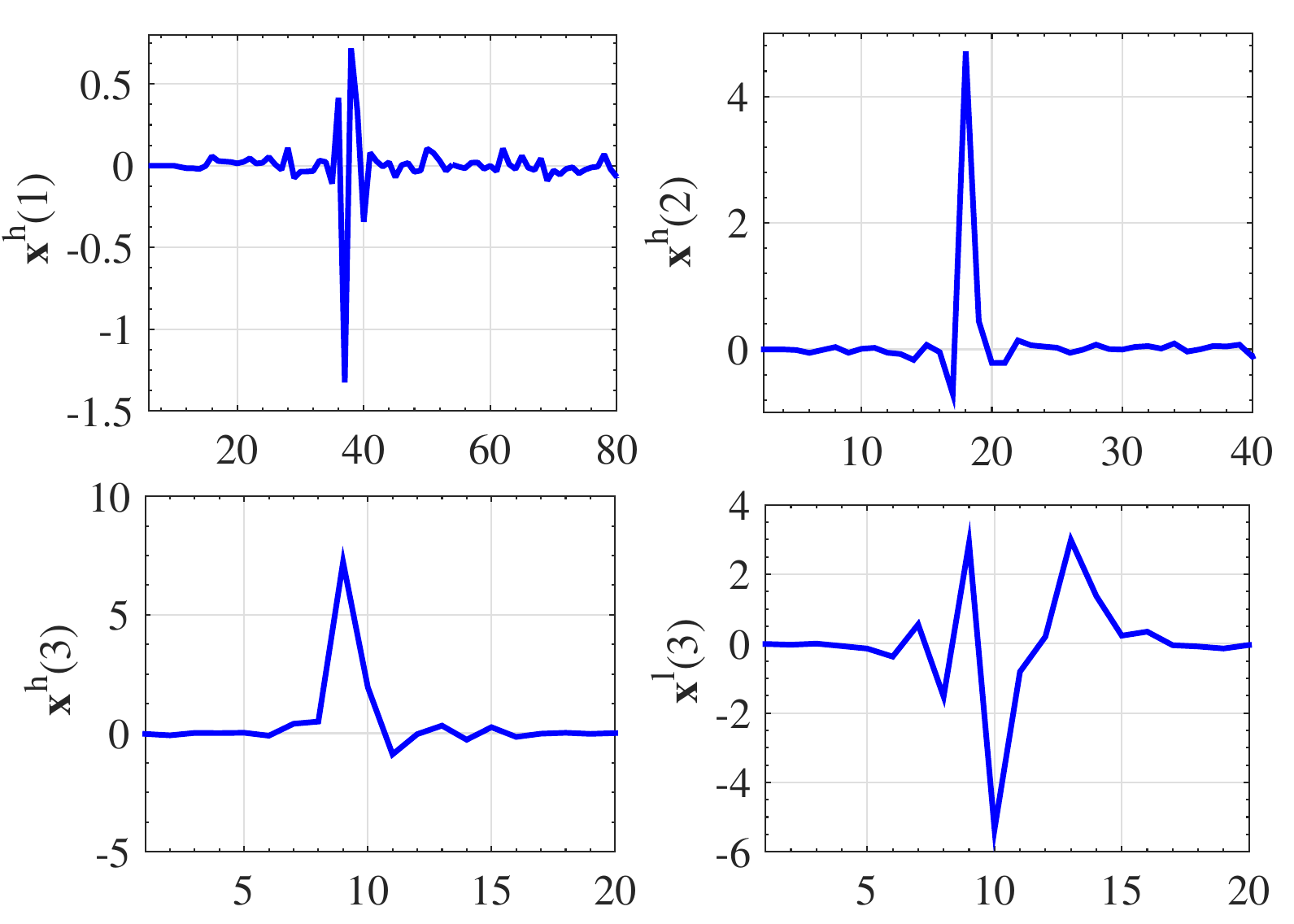}}
	\caption{Sub-series generated by the mWDN model.}\label{fig:wavelets}
\end{figure*}

Fig.~\ref{fig:wavelets} shows the outputs of mWDN layers in the mLSTM and RCF models fed with the two samples given in Fig.~\ref{fig:samples}, respectively. In Fig.~\ref{fig:wuxi_wavelet}, we plot the outputs of the first three layers in the mLSTM model as different sub-figures. As can be seen, from $\mathbf{x}^h(1)$ to $\mathbf{x}^l(3)$, the outputs of the middle layers correspond to the frequency components of the input series running from high to low. A similar phenomenon could be observed in Fig.~\ref{fig:ecg_wavelet}, where the outputs of the first three layers in the RCF model are presented. This phenomenon again indicates that the middle layers of mWDN inherit the frequency decomposition function of wavelet. Then here comes the problem: can we evaluate quantitatively what layer or which frequency of a time series is more important to the final output of the mWDN based models? If possible, this can provide valuable interpretability to our mWDN model.


\subsection{Importance Analysis}
We here introduce an importance analysis method for the proposed mWDN model, which aims to quantify the importance of each middle layer to the final output of the mWDN based models.

We denote the problem of time series classification/forecasting using a neural network model as
\begin{equation}
p = M(\mathbf{x}),
\end{equation}
where $M$ denotes the neural network, $\mathbf{x}$ denotes the input series, and $p$ is the prediction. Given a well-trained model $M$, if a small disturbance $\varepsilon$ to the $i$-th element $x_i\in \mathbf{x}$ can cause a large change to the output $p$, we say $M$ is sensitive to $x_i$. Therefore, the sensibility of the network $M$ to the $i$-th element $x_i$ of the input series is defined as the partial derivatives of $M(\mathbf{x})$ to $x_i$ as follows:
\begin{equation}\label{equ:sensibility_input}
S(x_i) = \left|\frac{\partial M({x}_i)}{\partial x_i}\right|
= \left|\lim_{\varepsilon \rightarrow 0} \frac{M({x}_i) - M({x}_i - \varepsilon)}{\varepsilon}\right|.
\end{equation}
Obviously, $S(x_i)$ is also a function of $x_i$ for a given model $M$. Given a training data set $\mathcal{X} = \{\tilde{\mathbf{x}}^1, \cdots, \tilde{\mathbf{x}}^j, \cdots, \tilde{\mathbf{x}}^J\}$ with $J$ training samples, the importance of the $i$-th element of the input series $\mathbf{x}$ to the model $M$ is defined as
\begin{equation}\label{equ:importance_input}
I(x_i) = \frac{1}{J} \sum_{j=1}^J S(\tilde{x}_i^j),
\end{equation}
where $\tilde{x}_i^j$ is the value of the $i$-th element in the $j$-th training sample.

The importance definition in Eq.~\eqref{equ:importance_input} can be extended to the middle layers in the mWDN model. Denoting $a$ as an output of a middle layer in mWDN, the neural network $M$ can be rewritten as
\begin{equation}
p = M(a(\mathbf{x})),
\end{equation}
and the sensibility of $M$ to $a$ is then defined as
\begin{equation}\label{equ:sensibility_a}
S_a(\mathbf{x}) = \left|\frac{\partial M(a(\mathbf{x}))}{\partial a(\mathbf{x})}\right|
= \left|\lim_{\varepsilon \rightarrow 0} \frac{M(a(\mathbf{x})) - M(a(\mathbf{x}) - \varepsilon)}{\varepsilon}\right|.
\end{equation}
Given a training data set $\mathcal{X}= \{\tilde{\mathbf{x}}^1, \cdots, \tilde{\mathbf{x}}^j, \cdots, \tilde{\mathbf{x}}^J\}$, the importance of $a$ {\it w.r.t.} $M$ is calculated as
\begin{equation}\label{equ:importance_a}
I(a) = \frac{1}{J}\sum_{j=1}^J S_a(\tilde{\mathbf{x}}^j).
\end{equation}
The calculation of $\frac{\partial M}{\partial \mathbf{x}_i}$ and $\frac{\partial M}{\partial a}$ in~Eq.~\eqref{equ:sensibility_input} and Eq.~\eqref{equ:sensibility_a} are given in the Appendix for concision. Eq.~\eqref{equ:importance_input} and Eq.~\eqref{equ:importance_a} respectively define the importance of a time-series element and an mWDN layer to an mWDN based model.

\begin{figure}[t]
	\centering
	\includegraphics[width=\columnwidth]{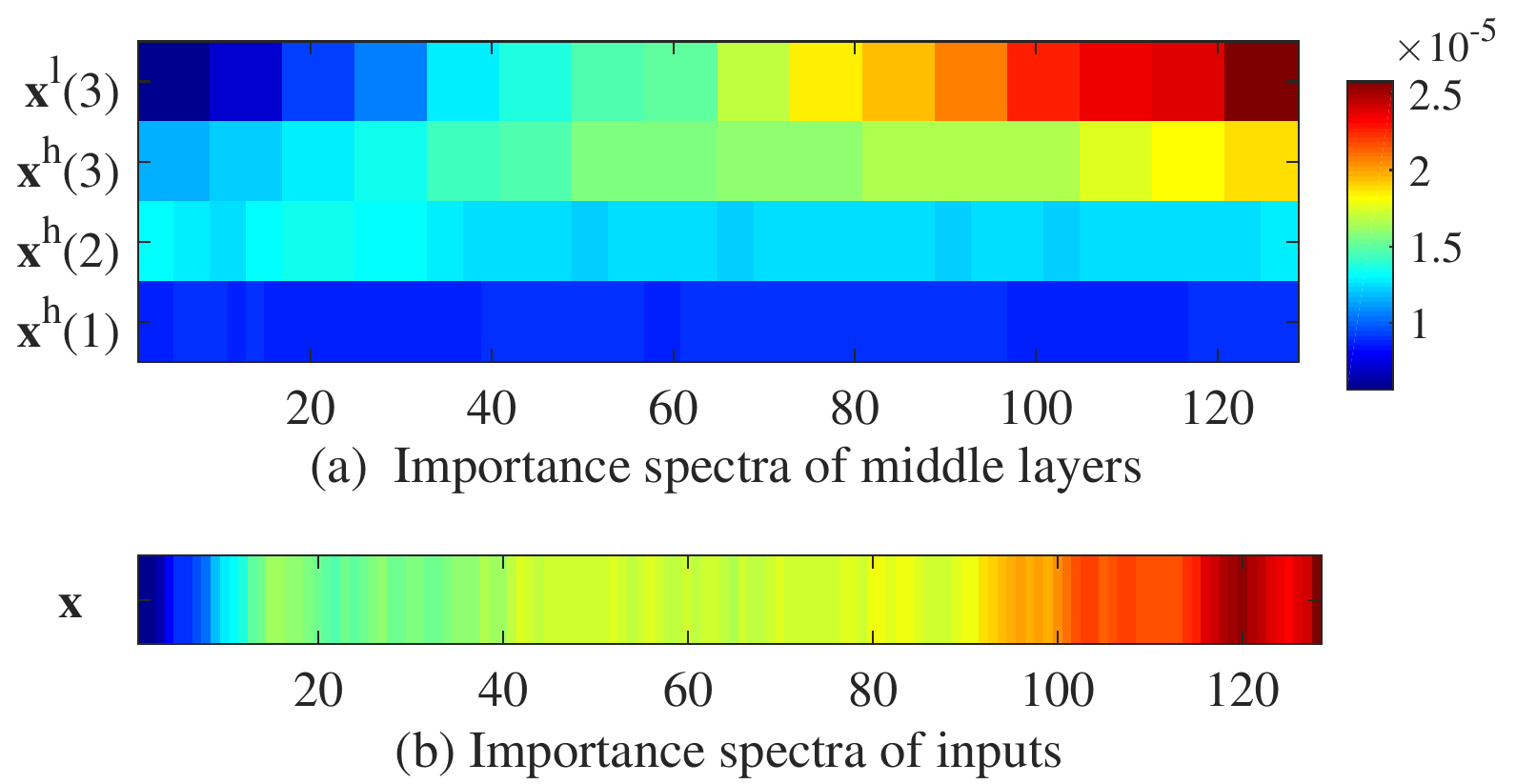}
	\caption{Importance spectra of mLSTM on \textit{WuxiCellPhone}.}\label{fig:importance_wuxi}
\end{figure}

\subsection{Experimental Results}
\begin{figure}[t]
	\centering
	\includegraphics[width=\columnwidth]{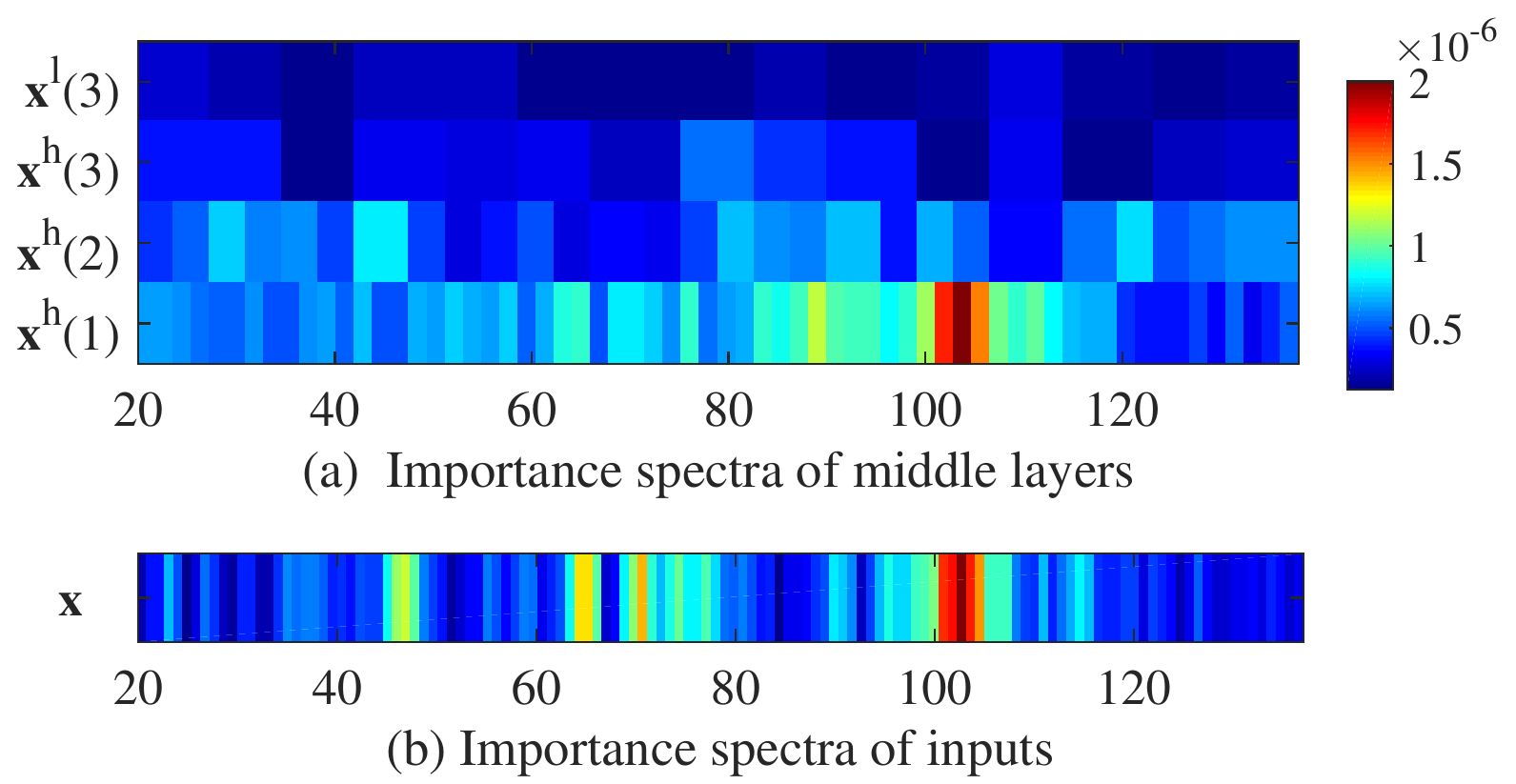}
	\caption{Importance spectra of RCF on \textit{ECGFiveDays}.}\label{fig:importance_ecg}
\end{figure}

Fig.~\ref{fig:importance_wuxi} and Fig.~\ref{fig:importance_ecg} shows the results of importance analysis. In Fig.~\ref{fig:importance_wuxi}, the mLSTM model trained on \textit{WuxiCellPhone} in Sect.~\ref{subsec:Forecasting} is used. Fig.~\ref{fig:importance_wuxi}(b) exhibits the importance spectrum of all the elements, where the x-axis denotes the increasing timestamps and the colors in spectrum denote the varying importance of the features: the redder, the more important. From the spectrum, we can see that the latest elements are more important than the older ones, which is quite reasonable in the scenario of time series forecasting and justifies the time value of information.

Fig.~\ref{fig:importance_wuxi}(a) exhibits the importance spectra of the middle layers listed from top to bottom in the increasing order of frequency. Note that for the sake of comparison, we resize the lengths of the outputs to the same. From the figure, we can observe that $i$) the lower frequency layers in the top are with higher importance, and $ii$) only the layers with higher importance exhibit the time value of the elements as in Fig.~\ref{fig:importance_wuxi}(b). These imply that the low frequency layers in mWDN are crucially important to the success of time series forecasting. This is not difficult to understand since the information captured by low frequency layers often characterizes the essential tendency of human activities and therefore is of great use to revealing the future.


Fig.~\ref{fig:importance_ecg} depicts the importance spectra of the RCF model trained on the \textit{ECGFiveDay} data set in Sect.~\ref{subsec:Classification}. As shown in Fig.~\ref{fig:importance_ecg}(b), the most important elements are located in the range from roughly 100 to 110 of the time axis, which is quite different from that in Fig.~\ref{fig:importance_wuxi}(b). To understand this, recall Fig.~\ref{fig:ecg_org} that this range corresponds to the T-Wave of electrocardiography, covering the period of the heart relaxing and preparing for the next contraction. It is generally believed that abnormalities in the T-Wave can indicate seriously impaired physiological functioning~\footnote{\url{https://en.m.wikipedia.org/wiki/T_wave}}. As a result, the elements describing T-Wave are more important to the classification task.

Fig.~\ref{fig:importance_ecg}(a) shows the importance spectra of middle layers, also listed from top to bottom in the increasing order of frequency. It is interesting that the phenomenon is opposite to the one in Fig.~\ref{fig:importance_wuxi}(a); that is, the layers in high frequency are more important to the classification task on \textit{ECGFiveDays}. To understand this, we should know that the general trends of ECG curves captured by low frequency layers are very similar for everyone, whereas the abnormal fluctuations captured by high frequency layers are the real distinguishable information for heart diseases identification. This also indicates the difference between a time-series classification task and the a time-series forecasting task.



{\bf Summary.} The experiments in this section demonstrate the interpretability advantage of the mWDN model stemming from the integration of wavelet decomposition and our proposed importance analysis method. It can also be regarded as an indepth exploration to solve the black box problem of deep learning.

\section{Related Works}

{\bf Time Series Classification (TSC).} The target of TSC is to assign a time series pattern to a specific category, {\it e.g.}, to identify a word based on series of voice signals. Traditional TSC methods could be classified into three major categories: distance based, feature based, and ensemble methods~\cite{MCNN}. Distance based methods predict the category of a time series by comparing the distances or similarities to other labeled series. The widely used TSC distances includes the Euclidean distance and dynamic time warping (DTW)~\cite{dtw}, and DTW with KNN classifier has been the state-of-the-art TSC method for a long time~\cite{keogh2005exact}. A defect of distance based TSC methods is the relatively high computational complexity. Feature based methods overcome this defect by training classifiers on deterministic features and category labels of time series. Traditional methods, however, usually depend on handcraft features as inputs, such as symbolic aggregate approximation and interval mean/deviation/slop~\cite{sax,deng2013time}. In recent years, automatic feature engineering was introduced to TSC, such as time series shapelets mining~\cite{shapelet3}, attention~\cite{qin2017dual} and deep learning based representative learning~\cite{langkvist2014review}. Our study also falls in this area but with frequency awareness. The well-known ensemble methods for TSC include PROP~\cite{prop}, COTE~\cite{COTE}, {\it etc.}, which aim to improve classification performance via knowledge integration. As reported by some latest works~\cite{MCNN,baselines}, however, existing ensemble methods are yet inferior to some distance based deep learning methods.

{\bf Time Series Forecasting (TSF).} TSF refers to predicting future values of a time series using past and present data, which is widely adopted in nearly all application domains~\cite{wang6,wang5}. A classic model is autoregressive integrated moving average (ARIMA)~\cite{arima}, with a great many variants, {\it e.g.}, ARIMA with explanatory variables (ARIMAX)~\cite{ARIMAX} and seasonal ARIMA (SARIMA)~\cite{SARIMA}, to meet the requirements of various applications. In recent years, a tendency of TSF research is to introduce supervised learning methods, such as support vector regression~\cite{OLWSVR} and deep neural networks~\cite{arima_nn}, for modeling complicated non-linear correlations between past and future states of time series. Two well-known deep neural network structures for TSF are recurrent neural networks (RNN)~\cite{rnn_tsf} and long short-term memory (LSTM)~\cite{lstm_tsf}. These indicate that an elaborate model design is crucially important for achieving excellent forecasting performance.

{\bf Frequency Analysis of Time Series.} Frequency analysis of time series data has been deeply studied by the signal processing community. Many classical methods, such as Discrete Wavelet Transform~\cite{mallat1989theory}, Discrete Fourier~\cite{dft}, and Z-Transform~\cite{z_transform}, have been proposed to analysis the frequency pattern of time series signals. In existing TSC/TSF applications, however, transforms are usually used as an independent step in data preprocessing~\cite{MCNN,waveletnn}, which have no interactions with model training and therefore might not be optimized for TSC/TSF tasks from a global view. In recent years, some research works, such as Clockwork RNN~\cite{cwrnn} and SFM~\cite{hu2017state}, begins to introduce the frequency analysis methodology into the deep learning framework. To our best knowledge, our study is among the very few works that embed wavelet time series transforms as a part of neural networks so as to achieve an end-to-end learning.


\section{Conclusions}

In this paper, we aim at building frequency-aware deep learning models for time series analysis. To this end, we first designed a novel wavelet-based network structure called mWDN for frequency learning of time series, which can then be seamlessly embedded into deep learning frameworks by making all parameters trainable. We further designed two deep learning models based on mWDN for time series classification and forecasting, respectively, and the extensive experiments on abundant real-world datasets demonstrated their superiority to state-of-the-art competitors. As a nice try for interpretable deep learning, we further propose an importance analysis method for identifying important factors for time series analysis, which in turn verifies the interpretability merit of mWDN.



\bibliographystyle{ACM-Reference-Format}
\bibliography{wavelet}

\section*{Appendix}
In a neural network model, the outputs of the layer $l$ are connected as the inputs of the layer $l+1$. According to the chain rule, the partial derivative of the model $M$ to middle layer outputs could be calculated layer-by-layer as
\begin{equation}\label{}
\frac{\partial M}{\partial a_i^{(l)}} = \sum_j \frac{\partial M}{\partial a_j^{(l+1)}} \frac{\partial a_j^{(l+1)}}{\partial a_i^{(l)}},
\end{equation}
where $a_i^{(l)}$ is the $i$-th output of the layer $l$. The proposed models contain types of layers: the convolutional, LSTM and fully connected layers, which are discussed below.

For convolutional layers, only 1D convolutional operation is used in our cases. The output of the layer $l$ is a matrix with the size of $L \times 1 \times C$, which is connected to neural matrix of the $l+1$-th with a convolutional kernel in the size of $k \times 1 \times C$. The partial derivative of $M$ to the $i^{th}$ output of the layer $l$ is calculated as
\begin{align*}
\frac{\partial M}{\partial a_{i}^{(l)}}
&= \sum_{n = 0}^{k - 1}  \frac{\partial M}{\partial a_{i-n}^{(l+1)}}\frac{\partial a_{i-n}^{(l+1)}}{\partial a_{i}^{(l)}}\\
&= \sum_{n = 0}^{k - 1}  \delta^{(l+1)}_{i - n} w_{n}^{(l+1)} f'\left(a_{i}^{(l)}\right),
\end{align*}
where $w_{n}$ denotes the $n$-th element of the convolutional kernel, $\delta^{(l)}_{i} = \frac{\partial M}{\partial a_{i}^{(l)}}$, and $f'\left(a_{i}^{(l)}\right)$ is the derivative of activation function.

For LSTM laysers, we denote the output of a LSTM unit in layer $l+1$ at time $t$ as
\begin{equation*}\label{}
a_i^{t,(l+1)} = f\left(b^{t,(l)}\right),
\end{equation*}
where $b^t(l)$ is calculated as
\begin{equation*}\label{}
b^{t,(l)} = \sum_i w_i^a a_i^{t,(l)} + \sum_i w_i^b b_i^{t-1,(l)} + \sum_i w_i^s s_i^{t-1,(l)}.
\end{equation*}
$s_i^{t-1,(l)}$ is the history state that is saved in the memory cell. Therefore, the partial derivative of $M$ to the $a_i^{t,(l)}$ is calculated as
\begin{align*}
\frac{\partial M}{\partial a_i^{(l)}} & = \sum_t \frac{\partial M}{\partial b^{t,(l)}}  \frac{\partial b^{t,(l)}}{\partial a_i^{t,(l)}} = \sum_t  \delta_i^{t,(l+1)} f'(b^{t,(l)}) \theta_i^{t,(l)},
\end{align*}
where $\theta_i^{t,(l)}$ is an equation as
\begin{equation*}
\theta_i^{t,(l)} = \left(w_i^a + w_i^b \frac{\partial b^{t+1,(l)}}{\partial a_i^{t+1,(l)}} + w_i^s \frac{\partial s^{t+1,(l)}}{\partial a_i^{t+1,(l)}} \right).
\end{equation*}
The derivative $\frac{\partial s^{t,(l)}}{\partial a_i^{t,(l)}}$ in the above equation is calculated as
\begin{equation*}\footnotesize
\frac{\partial s^{t,(l)}}{\partial a_i^{t,(l)}} = s^{t-1,(l)}\frac{\partial b^{t,(l)}}{\partial a_i^{t,(l)}} + \frac{\partial b^{t,(l)}}{\partial a_i^{t,(l)}} f(a_i^{t,(l)}) + b^{t,(l)} f'(a_i^{t,(l)}).
\end{equation*}

For fully connect layers, the output $a_i^{(l)} = f(w_i a_i^{(l-1)} + b)$. Then the partial derivative is equal to $$w_{i} f'(w_i a_{i}^{(l-1)} + b ).$$

\end{document}